\documentclass{ieeeaccess}
\usepackage{cite}
\usepackage{amsmath,amssymb,amsfonts}
\usepackage{graphicx}
\usepackage{textcomp}
\usepackage{subcaption}
\usepackage{bm}
\usepackage{algorithm}%
\usepackage{hyperref}
\usepackage{algpseudocode}%
\usepackage{float}
\usepackage{makecell}
\usepackage{longtable}
\usepackage{booktabs} 
\usepackage{multirow}
\usepackage{wrapfig} 
\usepackage{blindtext} 

\makeatletter
\AtBeginDocument{\DeclareMathVersion{bold}
\SetSymbolFont{operators}{bold}{T1}{times}{b}{n}
\SetSymbolFont{NewLetters}{bold}{T1}{times}{b}{it}
\SetMathAlphabet{\mathrm}{bold}{T1}{times}{b}{n}
\SetMathAlphabet{\mathit}{bold}{T1}{times}{b}{it}
\SetMathAlphabet{\mathbf}{bold}{T1}{times}{b}{n}
\SetMathAlphabet{\mathtt}{bold}{OT1}{pcr}{b}{n}
\SetSymbolFont{symbols}{bold}{OMS}{cmsy}{b}{n}
\renewcommand\boldmath{\@nomath\boldmath\mathversion{bold}}}
\makeatother

\def\BibTeX{{\rm B\kern-.05em{\sc i\kern-.025em b}\kern-.08em
    T\kern-.1667em\lower.7ex\hbox{E}\kern-.125emX}}

\begin{document}
\history{Date of publication xxxx 00, 0000, date of current version xxxx 00, 0000.}
\doi{10.1109/ACCESS.2024.0429000}

\title{Model Input-Output Configuration Search with Embedded Feature Selection for Sensor Time-series and Image Classification}
\author{\uppercase{Anh Tuan Hoang}\authorrefmark{1}\authorrefmark{2}, \IEEEmembership{Member, IEEE},
\uppercase{Zsolt János Viharos}\authorrefmark{1}\authorrefmark{3},
\IEEEmembership{Member, IEEE}}

\address[1]{HUN-REN Institute for Computer Science and Control, Center of Excellence in Production Informatics and Control (EPIC), Center of Excellence of the Hungarian Academy of Sciences (MTA), Budapest, 1111 Hungary}
\address[2]{Doctoral School of Informatics, Eötvös Loránd University (ELTE), Budapest, 1117 Hungary}
\address[3]{Faculty of Economics and Business, John von Neumann University, Kecskemét, 6000 Hungary}

\tfootnote{This research has been supported by the European Union project RRF-2.3.1-21-2022-00004 within the framework of the Artificial Intelligence National Laboratory and by the TKP2021-NKTA-01  NRDIO grant on "Research on cooperative production and logistics systems to support a competitive and sustainable economy". On behalf of Project "Comprehensive Testing of Machine Learning Algorithms" we thank for the usage of HUN-REN Cloud (https://science-cloud.hu/) that significantly helped us achieving the results published in this paper.}

\markboth
{Author \headeretal: Preparation of Papers for IEEE TRANSACTIONS and JOURNALS}
{Author \headeretal: Preparation of Papers for IEEE TRANSACTIONS and JOURNALS}

\corresp{Corresponding author: Anh Tuan Hoang (e-mail: hoang@sztaki.hu).}

\begin{abstract}
Machine learning is a powerful tool for extracting valuable information and making various predictions from diverse datasets. Traditional machine learning algorithms rely on well-defined input and output variables; however, there are scenarios where the separation between the input and output variables and the underlying, associated input and output layers of the model are unknown. Feature Selection (FS) and Neural Architecture Search (NAS) have emerged as promising solutions in such scenarios. This paper proposes MICS-EFS, a Model Input-Output Configuration Search with Embedded Feature Selection. The methodology explores internal dependencies in the complete input parameter space for classification tasks involving both 1D sensor time-series and 2D image data. MICS-EFS employs a modified encoder-decoder model and the Sequential Forward Search (SFS) algorithm, combining input-output configuration search with embedded feature selection. Experimental results demonstrate MICS-EFS’s superior performance compared to other FS algorithms. Across all tested datasets, MICS-EFS delivered an average accuracy improvement of 1.5\% over baseline models, with the accuracy gains ranging from 0.5\% to 5.9\%. Moreover, the algorithm reduced feature dimensionality to just 2–5\% of the original data, significantly enhancing computational efficiency. These results highlight the potential of MICS-EFS to improve model accuracy and efficiency in various machine learning tasks. Furthermore, the proposed method has been validated in a real-world industrial application focused on machining processes, underscoring its effectiveness and practicality in addressing complex input-output challenges.
\end{abstract}


\titlepgskip=-21pt

\maketitle

\section{Introduction}
\label{sec:introduction}
\begin{figure*}[h!]
    \centering
    \begin{subfigure}{0.4\textwidth}
        \centering
        \includegraphics[width=\linewidth]{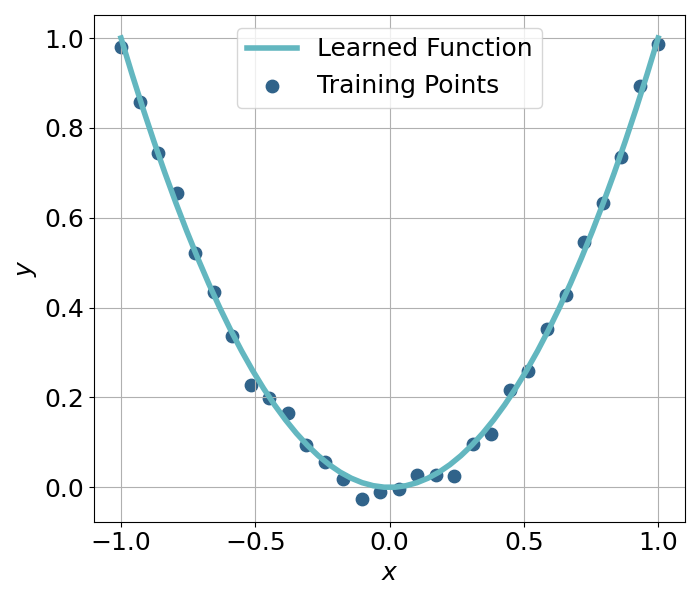}
        \caption{Well-chosen configuration}
        \label{fig:subfig_a}
    \end{subfigure}
    \hspace*{1cm}
    \begin{subfigure}{0.4\textwidth}
        \centering
        \includegraphics[width=\linewidth]{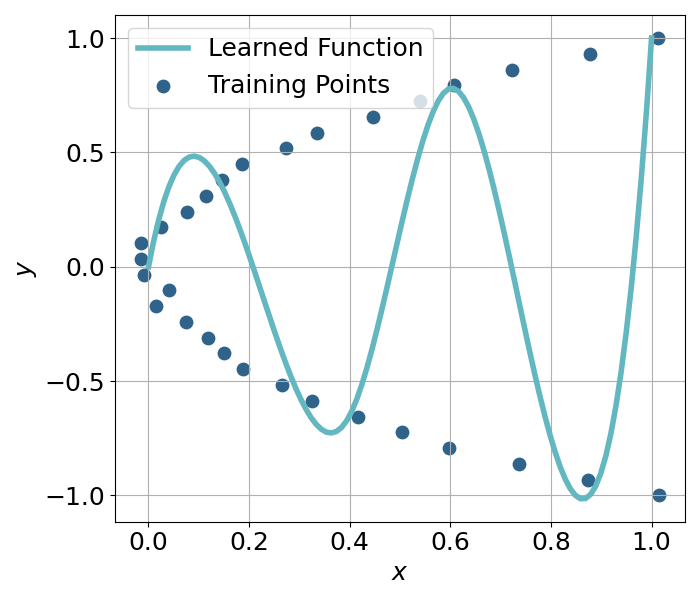}
        \caption{Poorly-chosen configuration}
        \label{fig:subfig_b}
    \end{subfigure}
    \hspace*{0.5cm}
    \caption{Given parameters: $x$ and $x^2$. a) The relationship $x \xrightarrow{} x^2$ (with x as input and $x^2$ as output) is attempted to be learned. This configuration is well-chosen, allowing the function to be effectively learned. b) The relationship $x^2 \xrightarrow{} x$, (with $x^2$ as input and $x$ as output), is attempted to be learned. This configuration is poorly chosen, resulting in inaccurate learning of the function.}
    \label{fig:x2}
\end{figure*}

Machine learning has emerged as a powerful tool for extracting valuable information and making predictions from complex and diverse datasets. Traditional machine learning algorithms rely on well-defined input and output variables, which provide clear guidelines for model training and inference, enabling them to learn patterns and relationships between them. However, there are scenarios where the distinction between the input and output variables and the underlying, associated architecture of the related model, are unknown or not readily available \cite{willems1997introduction} \cite{willems2007behavioral}. In such cases, the challenge lies in developing machine learning techniques that can autonomously discover the optimal input-output mapping, the related neural architecture and also the relevant features of the model, all in one cohesive algorithm that brings them together seamlessly. This has led to the emergence of Feature Selection (FS) \cite{chandrashekar2014survey} \cite{akinola2022multiclass} and Neural Architecture Search (NAS) \cite{elsken2019neural} as promising approaches to address these challenges. A simple example is shown in Figure \ref{fig:x2}, where the input-output configurations of the relationship between $x$ and $x^2$ illustrate the importance of selecting the optimal configuration.

Another example can be found in real-world applications, such as the modeling of manufacturing operations and machining processes \cite{viharos1999automatic} \cite{viharos1999selection} \cite{viharos2016optimal}. In these scenarios, the same parameter space can support multiple input-output parameter configurations depending on the task, with each configuration emphasizing different aspects of the process.

In the era of big data, where datasets are growing exponentially in both size and dimensionality, the need for efficient feature selection techniques has become prevalent in machine learning and data analysis \cite{chandrashekar2014survey}. Feature selection plays a pivotal role in addressing the challenges posed by high-dimensional datasets, such as computational complexity, overfitting and the curse of dimensionality \cite{zebari2020comprehensive}. It aims to identify the most relevant subset of features that are crucial for accurate and robust predictive modelling. The goal of feature selection is twofold: to enhance the performance of machine learning models and to improve their interpretability. By selecting a subset of discriminative features, feature selection techniques aim to reduce noise, irrelevant information and redundant features, which can lead to improved model generalization and reduced computational cost. In many real-world applications, understanding the factors that drive model predictions is essential for building trust, and explaining decisions \cite{eljialy2024novel} \cite{majidpour2024nsga}.

NAS is a field within machine learning that aims to automate the design of neural network architectures. NAS has primarily focused on optimizing the network structure, such as the number of layers, their connectivity and hyperparameters \cite{elsken2019neural}. It involves the use of algorithms to search and discover optimal network architectures for a given task, such as image classification \cite{8579005} \cite{real2019aging}, object detection \cite{8579005} and semantic segmentation \cite{chen2018searching}. NAS methods explore a vast space of possible architectures by iteratively evaluating and comparing different configurations. These methods often use reinforcement learning \cite{zoph2017neural} \cite{baker2017designing}, evolutionary algorithms \cite{real2017large} \cite{real2019aging} or gradient descent method \cite{liu2018darts} to guide the search process. However, one limitation in NAS is the lack of support for determining the input-output parameter configuration, which is also an important, but more rare aspect of architecture identification. This means that while NAS can automatically design the structure of a neural network, determining the appropriate input and output parameter configuration still requires human intervention and domain expertise. The determination of input configuration is closely connected to the feature selection topic, as it implicitly selects feature components with the related layers that capture relevant information from the data. Therefore, addressing the input-output parameter configuration search and its relationship to feature selection is crucial for further advancements which is currently a mainly unexplored area.

The integration of input-output configuration search with feature selection presents several advantages. First, input-output configuration search can effectively explore a vast space of potential input feature combinations, selecting only the most informative features for the given task. This automated process reduces the computational complexity and human effort associated with manual feature selection. Moreover, the combined method can capture complex interactions and dependencies between features, leading to more accurate and robust models. In this research, a comprehensive description of the algorithm called MICS-EFS: Model Input-Output Configuration Search with Embedded Feature Selection is proposed, designed to work with both sensor time-series data and image data. 

One of the main challenges addressed by the MICS-EFS algorithm is the lack of comprehensive integration between feature selection and input-output configuration search. Traditional methods often treat feature selection and neural architecture optimization as separate processes, resulting in suboptimal exploration of the dependencies among input and output parameters. Another significant challenge lies in handling complex datasets with high-dimensional features, where redundant or irrelevant data can obscure critical patterns. Furthermore, existing NAS techniques typically fix the input-output configuration, limiting their adaptability to scenarios where these mappings are not predefined. MICS-EFS aims to bridge these gaps by introducing a unified framework that autonomously discovers optimal input-output configurations while simultaneously identifying the most relevant features, ensuring enhanced model performance and efficiency.

The key contributions and novelties of this work are as follows: 

\begin{itemize}
  \item[--] \textit{Unified Framework:} A novel algorithm called MICS-EFS is introduced, combining input-output configuration exploration with feature selection in a single cohesive framework. 
\vspace{0.15cm}
  \item[--] \textit{Modified Autoencoder Approach:} A modified autoencoder model is proposed, capable of handling arbitrary input-output configurations for both 1D sensor time-series and 2D image data.
\vspace{0.15cm}
  \item[--] \textit{Integrated Feature Selection:} An embedded feature selection mechanism is incorporated, allowing the simultaneous identification of optimal features and input-output dependencies.
\vspace{0.15cm}
  \item[--] \textit{Practical Validation:} Comprehensive validation is conducted using diverse datasets, including industrial machining processes, showcasing the robustness and adaptability of the proposed algorithm in real-world applications.
\vspace{0.15cm}
  \item[--] \textit{Enhanced Performance:} The method consistently outperforms state-of-the-art solutions, achieving superior classification accuracy and reducing computational requirements by identifying the most informative features with only 2–5\% of the original data. 
\vspace{0.15cm}
  \item[--] \textit{Open Access Resource:} The algorithm's implementation is made publicly available, encouraging reproducibility and further research in this area.
  \end{itemize}

The following sections cover the related scientific works, the scientific motivation, introduction of the proposed MICS-EFS algorithm, experimental analysis and results, and conclusions with future works

\section{Related scientific works}
\label{sec:related}
In recent years, there is an explosive growth of the scientific literature on machine learning research, however it has seen limited progress in addressing the challenges of establishing mappings between unknown input and output variables and their underlying neural architecture. This section aims to offer a thorough literature review, emphasizing the integration of Feature Selection and Neural Architecture Search to confront these challenges. The research also involves domains, including system theory, which delves into the functioning of systems and their responses to different inputs \cite{willems1997introduction} \cite{willems2007behavioral} \cite{baillieul1999mathematical}. As it covers various domains, the research primarily focuses on Input-Output Configuration Search, Feature Selection and Nerual Architecture Search.

\subsection{Input-Output Configuration Search}
\begin{figure}[h]
     \centering
         \includegraphics[width=0.48\textwidth]{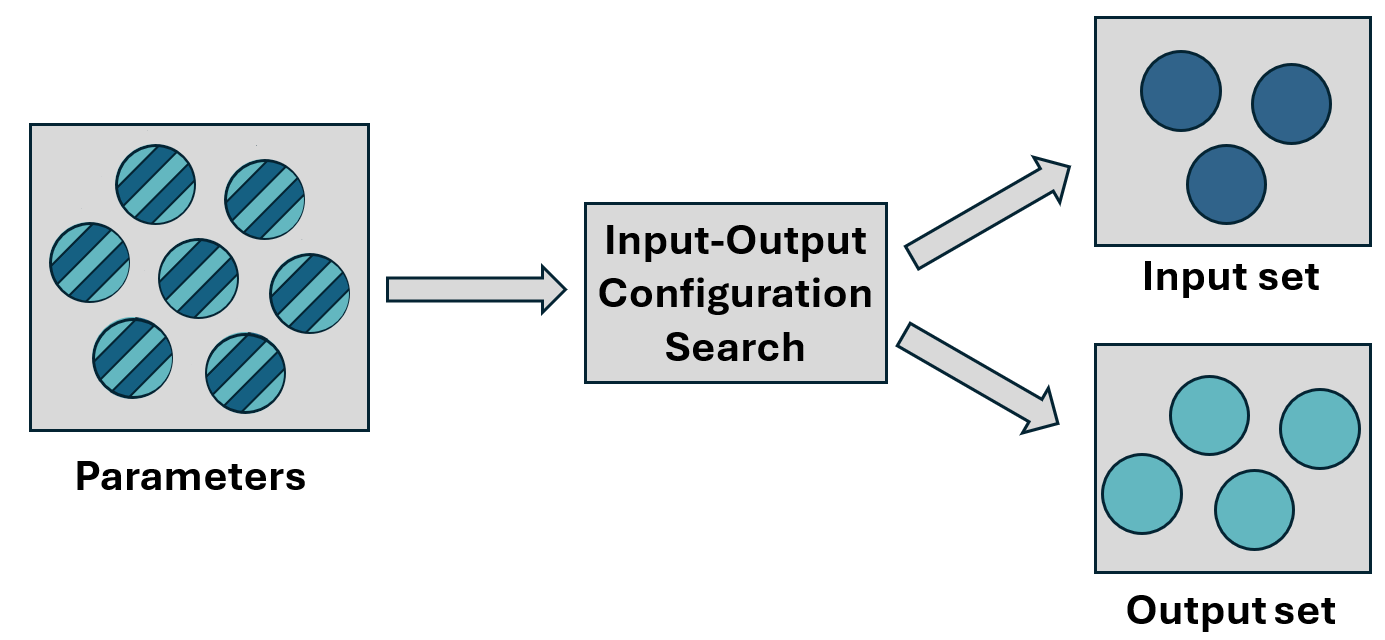}
         \caption{The concept of the Input-Output Configuration Search. The input-output properties of elements within the parameter space are initially unknown. Through this search process, the parameters are separated into two distinct sets, identifying which parameters behave more as inputs or outputs.} 
         \label{fig:concept}
\end{figure}

Configuration Search involves determining the optimal configuration of input and output parameters to achieve optimal performance and functionality in a model or system \cite{willems1997introduction} \cite{willems2007behavioral} (Figure \ref{fig:concept}).
Viharos et al. \cite{viharos1999automatic} \cite{monostori2001hybrid} introduced input-output configuration search algorithms, which have been applied to different learning models. They addressed the problem of automatic input-output configuration and generation of shallow neural network based process models with special emphasis on the modelling of production chains. Combined use of Sequential Forward Search (SFS), ANN learning and Simulated Annealing (SA) was proposed for the determination and application of general process models that meet the accuracy requirements of different tasks.

The applicability and extensibility of the concept were further introduced for ANFIS Neuro-Fuzzy Systems \cite{viharos2016optimal} \cite{viharos2015survey} and Support Vector Machines (SVMs) \cite{viharos2011support}, as they can be used to define the general model for a given problem. They identified the optimal input-output configuration of the system model, which implements the most accurate estimation that can be interpreted under the given conditions and explores the \textit{maximum of dependencies} among the related system parameters.

As a further extension, the sub-model decomposition algorithm has been developed, as an efficient method for automatically decomposing a complex system into several smaller interconnected sub-models and has been successfully applied in practice \cite{viharos2005automatic} \cite{viharos2007automatic}. It also enables more efficient training, as smaller sub-models can be trained independently, potentially reducing computational costs. Additionally, sub-model decomposition can enhance generalization, as each sub-model can learn specialized features and representations. 

\subsection{Feature Selection}
Feature Selection is the process of selecting a subset of relevant features from a larger set of available features to improve the performance and efficiency of a machine learning model.  There are several distinct approaches to feature selection, each with its own merits and considerations. The three main types of feature selection methods are filter, wrapper, and embedded techniques.

Filter methods involve assessing the relevance of features independently of any specific model \cite{guyon2003introduction}. These methods often rely on statistical metrics such as correlation, chi-squared tests, or mutual information to determine the relationship between each feature and the target variable. Wrapper methods, on the other hand, incorporate the machine learning model itself into the feature selection process \cite{kohavi1997wrappers}. These methods evaluate subsets of features by training and testing the model iteratively.  Although wrapper methods can provide more accurate feature subsets tailored to a specific model, they tend to be computationally intensive due to their reliance on the model's performance as a criterion. Embedded methods merge feature selection with the model training process \cite{wang2015embedded}. These techniques select features while the model is being trained, ensuring that the selected features are optimized for the model's performance.

Recently, several autoencoder-based \cite{autoencoder} feature selection methods have been developed. These methods leverage the power of autoencoders to learn compact representations of input data while simultaneously identifying the most informative features for the given task. Each FS method employs various techniques and architectures tailored to specific feature selection objectives.
Han et al. \cite{han2018autoencoder} proposed an unsupervised feature selection method that employs a self-representation autoencoder model and weights importance assigned to each feature. By simultaneously minimizing the reconstruction error and implementing group sparsity regularization, a subset of relevant features is obtained. Another method, called AgnoS was proposed by Doquet et al. \cite{doquet2020agnostic}, which employs an improved learning criterion to identify the initial features that are crucial for approximating all other the features. Balin et al. \cite{balin2019concrete} propose a novel approach called Concrete Autoencoder that utilizes a concrete selector layer as the encoder and a standard neural network as the decoder. By gradually decreasing the so called temperature of the concrete selector layer during training, the model learns a user-specified number of discrete features. A different approach was taken by Wu et al. \cite{wu2021fractal}, who introduced fractal autoencoders (FAE), that utilizes a neural network to precisely identify subset of features through global representability exploration, local diversity excavation, and one-to-one scoring layer. Following them, QuickSelection (QS) was presented by Atashgahi et al. \cite{atashgahi2022quick}. QS introduces the concept of neuron strength in sparse neural networks as a criterion for assessing feature importance. This criterion, combined with sparsely connected denoising autoencoders trained using the sparse evolutionary training procedure, allows the simultaneous determination of the importance of all input features. Finally, a similar solution was devised by Zang et al. \cite{zang2023udrn}, who developed Unified Dimensional Reduction Network (UDRN), a unified framework that seamlessly integrates Feature Selection (FS) and Feature Projection (FP) tasks. To achieve this integration, a novel network framework is devised, comprising a stacked feature selection network and a feature projection network, enabling separate execution of FS and FP tasks. 

\subsection{Neural Architecture Search}
Neural architecture search (NAS) \cite{elsken2019neural} has emerged as a promising field in the domain of deep learning, with the goal of automating the process of designing effective and efficient neural network architectures. NAS has gained significant attention in recent years due to its potential to alleviate the manual design process and improve the performance of deep learning models across various tasks.

One of the pioneering works in NAS is the Neural Architecture Search with Reinforcement Learning introduced by Zoph et al. \cite{zoph2017neural}. Their method utilizes a Recurrent Neural Network (RNN) as a controller that generates a sequence of architectural decisions. These decisions are then used to construct and evaluate neural network architectures. The controller is trained using Reinforcement Learning (RL), where the reward is based on the performance of the generated architectures. 
Following them, one notable approach is presented by Real et al. \cite{real2017large}\cite{real2019aging}, that neuro-evolution can construct large, accurate networks for challenging image classification benchmarks. It is capable of achieving this, starting from trivial initial conditions and searching a vast space. Importantly, the neuro-evolution process requires no experimenter participation and yields fully trained models. 

Another significant advancement in NAS is the inclusion of gradient-based methods. These methods leverage the gradient information of the validation loss with respect to the architectural parameters to guide the search process. Liu et al. proposed the DARTS (Differentiable Architecture Search) framework \cite{liu2018darts}, which formulates the search as a differentiable optimization problem. DARTS relaxes the discrete search space to a continuous space and uses a continuous relaxation of the architecture to obtain gradients. To further enhance the performance and address specific challenges, several variants of DARTS have been proposed, each introducing unique modifications and ideas. DARTS+ \cite{liang2019darts+} extends the original framework by introducing continuous relaxation at both macro and micro levels, allowing for finer control over information flow. GDAS (Gradient-based Differentiable Architecture Search) \cite{dong2019search} modifies DARTS with gradient-based sampling, improving efficiency in architecture selection. ProxylessNAS \cite{cai2018proxylessnas} introduces weight-sharing strategies to reduce search costs, while PC-DARTS \cite{xu2020pcdarts} implements progressive search for improved stability. Additionally, SNAS (Stochastic Neural Architecture Search) \cite{xie2018snas} incorporates stochastic sampling using Gumbel-Softmax, reducing computational overhead. Also, SeqNAS \cite{udovichenko2024seqnas} is  specifically designed for event sequence classification.

\section{Motivation}
\label{sec:motivation}
The research on novel NAS algorithms for ANNs is an active and also a promising future research field, as these methods already surpass popular and manually designed ANN architectures. The actual NAS algorithms concentrate only on the "\textit{internal}" structure of deep neural network models, as, e.g., the very famous DARTS algorithm and its variants do. "\textit{Internal}" means that these search algorithms consider the model structure "\textit{between}" the predetermined input and output variables but do not involve them in the architecture search process: inputs and outputs are considered fixed. \textit{The classical feature selection methodology goes beyond this concept as it already selects the rather relevant input variables, consequently, these parameters are not fixed anymore.} However, feature selection solutions do not modify completely the input-output structure of the models e.g., they do not select variables among the output parameters, consequently, not a comprehensive exploration of dependencies among all model parameters is given by feature selection.

The proposed algorithm realises a two-level approach: first, its main aim is to find "\textit{all}" dependencies among the input parameters through an integrated feature selection solution and in the next stage it exploits these explored dependencies in order to solve the original assignment. In this research, the superiorities of MICS-EFS are described enabling its wide applications in various practical and scientific fields.

\section{MICS-EFS: The proposed, novel method}
\label{sec:methodology}

The primary objective of this section is to introduce MICS-EFS algorithm, a novel Model Input-Output Configuration Search with Embedded Feature Selection. The MICS-EFS methodology focuses on exploring the dependencies given inside the input space, like image and signal time-series sources to determine suitable inputs and outputs for enhancing the classification/estimation performance of the model. 

\subsection{The concept of MICS-EFS}
\label{sec:concept}
In this section, a detailed explanation about the main concept behind the proposed MICS-EFS method will be provided. The method comprises three major components:

\begin{enumerate}
\vspace{0.15cm}
  \item \textit{Measurement:} measure the dependencies of an arbitrarily given input-output configuration. Here, this evaluation yields a numerical result known as the reconstruction cost, which reflects the accuracy of reconstructing the outputs from the given input to capture the underlying dependencies within the data.
\vspace{0.15cm}
  \item \textit{Search algorithm:} focuses on determining the optimal input-output configuration of the model by a search algorithm. It serves as an effective and ideal alternative approach to feature selection.
\vspace{0.15cm}
  \item \textit{Modeling methodology:} describes the implementation of a new evaluation procedure that utilizes the reconstructed data generated by the proposed method.
\end{enumerate}

\vspace{0.1cm}
\noindent By delving into these components, a comprehensive understanding of the MICS-EFS method and its functioning can be gained.

\subsubsection{Measurement: Exploration of I/O configuration}
\label{sec:input-output}
The proposed MICS-EFS method is based on a modified version of the traditional autoencoder model \cite{autoencoder}, specifically designed to explore the relationships for an arbitrarily given input-output configuration. The traditional autoencoder is a form of artificial neural network that learns efficient data encoding in an unsupervised manner, it comprises an encoder and a decoder component. The encoder maps the input data ($x$) to a lower-dimensional space Z, while the decoder reconstructs the input data by minimizing a predefined loss function ($L$). Here, the Z block represents the latent space or the compressed representation of the input data that captures the most significant features and dependencies. Formally, the objective is to learn the encoder function $f: \mathbb{R}^n \mapsto \mathbb{R}^p$ and the decoder function $g: \mathbb{R}^p \mapsto \mathbb{R}^n$, satisfying 

\begin{equation}\label{eq1}
argmin_{f, g}E[L(x, g\circ f(x))]
\end{equation}

\noindent , where $E$ is the expectation over the distribution of $x$, and the loss function often being the $\ell_2$ norm, although alternative measures exist \cite{viharos2022self}. Generally, the encoder and decoder are implemented using fully connected layers within the artificial neural network framework.

\begin{figure}[h]
     \centering
         \includegraphics[width=0.5\textwidth]{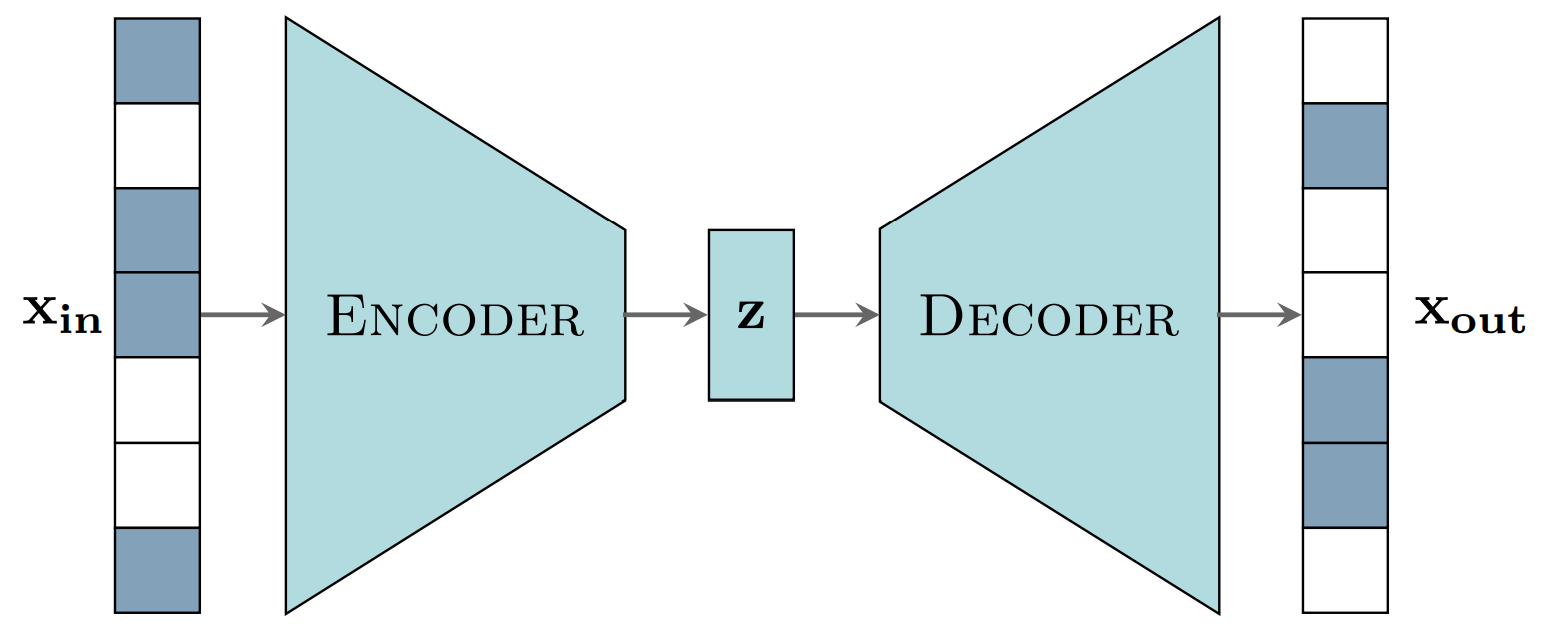}
         \caption{Modified adaptation of the classic autoencoder model with input $x_{in}$ and output $x_{out}$. Blue color indicates available and utilized, white color indicates missing (ignored) variables.} 
         \label{fig:autoencoder}
\end{figure}

The proposed MICS-EFS method employs a classic encoder-decoder model with adaptations tailored to explore the relationships between arbitrarily defined input and output sets. Unlike conventional autoencoders that reconstruct the original input, this approach partitions the data into distinct input and output subsets, namely the input set $x_{in}$ and the output set $x_{out}$, where $x_{in} \cup x_{out} = x$. While the underlying architecture remains largely standard, the methodology diverges by dynamically altering the input-output mapping, making it suitable for discovering intricate parameter dependencies in both 1D and 2D datasets. This adjustment enables the modified autoencoder to learn a mapping between an arbitrary given input-output configurations (see Figure \ref{fig:autoencoder}).
\vspace{0.5cm}

\begin{figure*}[h!]
\centering
\begin{subfigure}{\textwidth} 
\centering
\includegraphics[width=\linewidth]{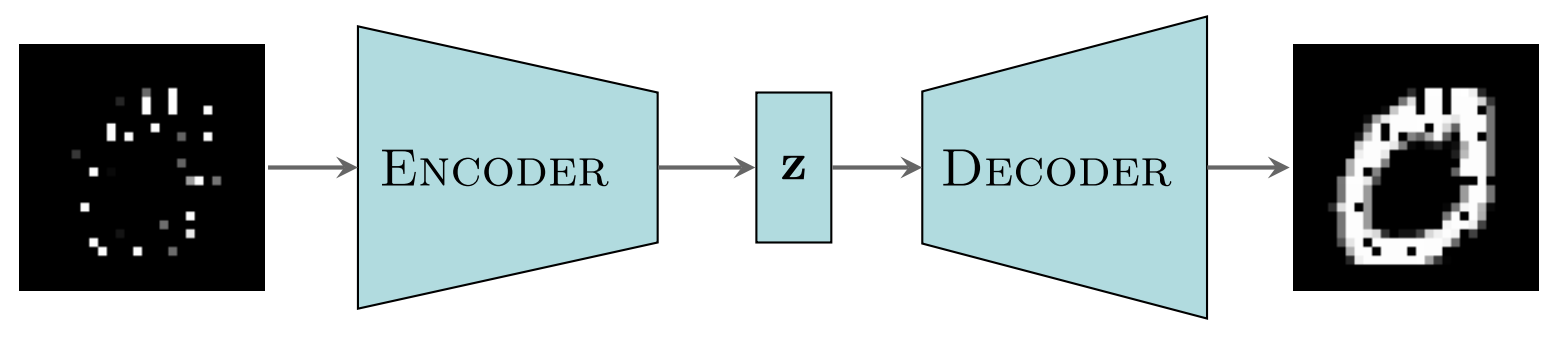}
\caption
{\centering 2D CNN}
\label{fig:autoencoder2a}

\end{subfigure}

\begin{subfigure}{\textwidth}
\centering
\includegraphics[width=\linewidth]{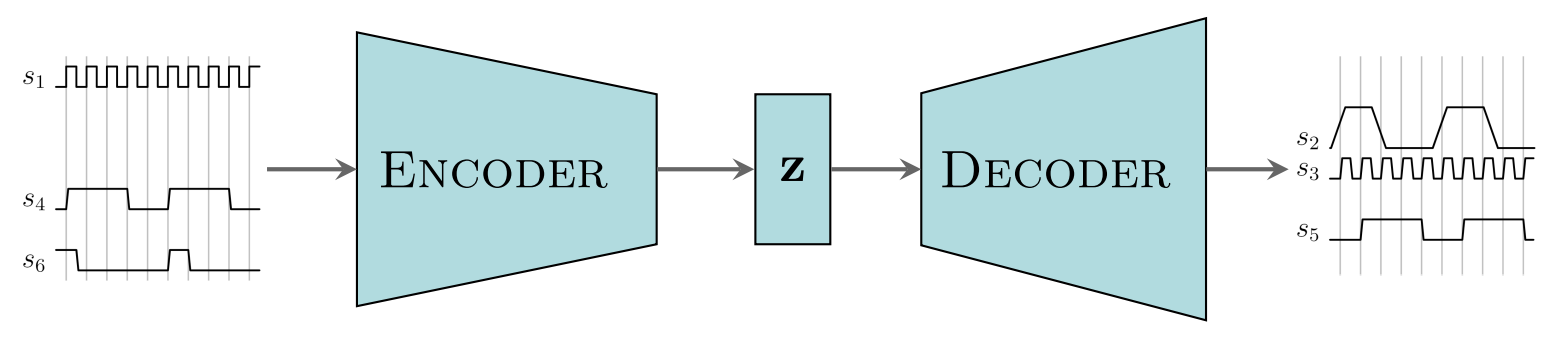}
\caption{\centering 1D CNN}
\label{fig:autoencoder2b}

\end{subfigure}

\caption
{\centering Modified adaptations of the classic convolutional autoencoder model with input $x_{in}$ and output $x_{out}$ for image (upper) and sensor time-series data (lower).}
\label{fig:autoencoder2}
\end{figure*}

Based on the new concept, Equation \ref{eq1} is modified as follows:

\begin{equation}\label{eq2}
argmin_{f, g}E[L(x_{out}, g\circ f(x_{in}))]
\end{equation}

\noindent The specialization of the proposed method is as follows: the proposed method can be extended to use convolutional autoencoders instead of standard ones. In a convolutional autoencoder, convolutional layers are utilized in the encoder part for dimensionality reduction, while transposed convolution layers are employed in the decoder part to restore the original data resolution. Consequently, the model can accommodate both 1D and 2D convolutional layers, or even in more dimensionality, enabling the processing of sersor time-series, image and multimodal data as well (Figure \ref{fig:autoencoder2}).

\subsubsection{Search algorithm: Sequential Forward Search} 
\label{sec:input-output2}

The primary objective of the MICS-EFS algorithm is to automatically determine the optimal input-output configuration that maximizes the estimation capability of the model. In Section \ref{sec:input-output},  internal dependencies using arbitrarily chosen input-output configurations are defined. Now, the focus shifts to the search for an optimal input-output configuration. To achieve this, a search algorithm is employed, which is another fundamental components of the method.

\begin{figure}[h]
     \centering
         \includegraphics[width=0.5\textwidth]{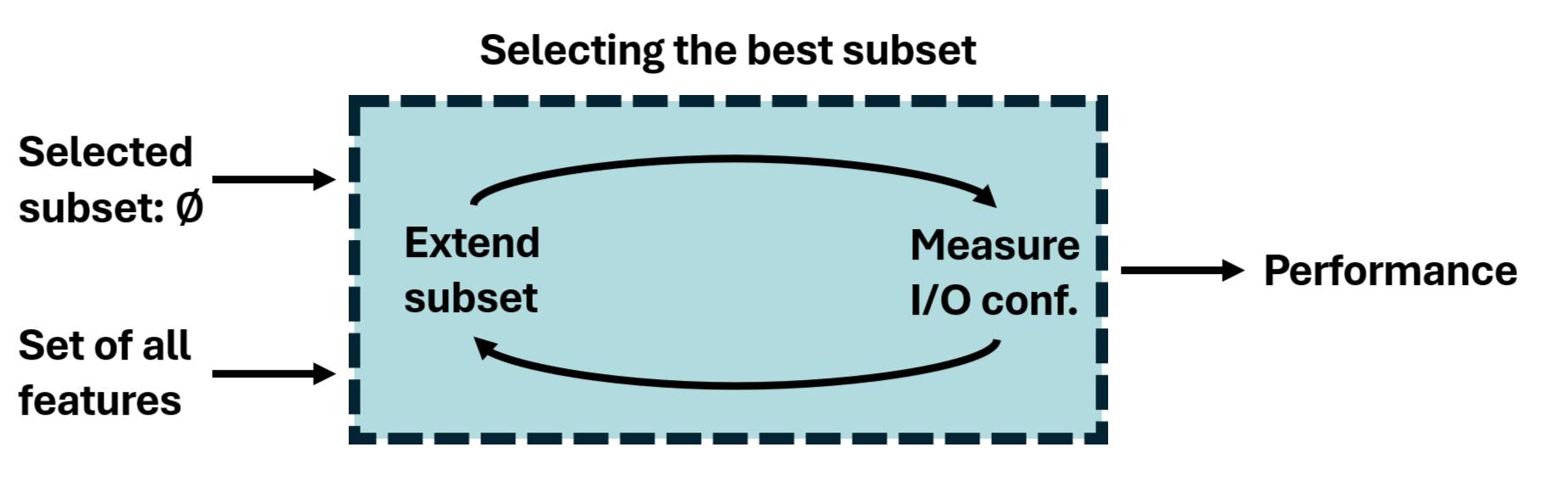}
         \caption{The process of the Sequential Forward Search algorithm. SFS iteratively builds an input feature subset by adding one feature at each iteration, selecting the one that has optimal input-output dependency, until a stopping criterion is met.} 
         \label{fig:search}
\end{figure}

During the research, Sequential Forward Search (SFS) was utilized as the chosen search algorithm. SFS is preferred for its simplicity and transparency, providing a clear understanding of the search process. However, it is important to note that potentially more efficient search algorithms could also be used, though exploring these alternatives falls outside the scope of this research. SFS operates by iteratively building an input feature subset by adding one input feature at a time based on a predefined criterion, typically using a performance measure such as accuracy or error rate. At each step, SFS evaluates the performance of the model with the current input feature subset and selects the input feature that improves the model's performance the most. This process continues until a stopping criterion is met, such as reaching a desired number of features or observing a decrease in model performance (Figure \ref{fig:search}).  

While SFS is favored for its simplicity and interpretability, it presents notable limitations that may impact its performance in certain scenarios. Firstly, SFS operates in a greedy manner, evaluating and adding features sequentially based on a predefined criterion. This approach, while efficient, inherently risks settling on locally optimal solutions rather than discovering the global optimum, particularly in cases involving complex or highly non-linear dependencies among features. Moreover, the algorithm's sensitivity to the order of feature addition can lead to suboptimal results if early choices are not representative of the broader dataset characteristics. For instance, a poorly chosen initial feature might constrain the search trajectory, resulting in an overall less informative feature subset. To mitigate these limitations, potential enhancements could involve the use of alternative search strategies, such as backward elimination, hybrid approaches, or even stochastic algorithms like genetic algorithms or simulated annealing, which offer a broader exploration of the feature space.

Within the MICS-EFS algorithm, there is an embedded feature selection mechanism that operates silently while searching for the optimal input-output configuration. This embedded feature selection algorithm identifies a subset of relevant features and, in the case of SFS, also provides a feature ranking. By employing this algorithm, the simultaneous search for the optimal input-output configuration is conducted while implicitly performing feature selection. This integration of 

\begin{figure*}[h!]
    \centering
        \includegraphics[width=0.75\linewidth]{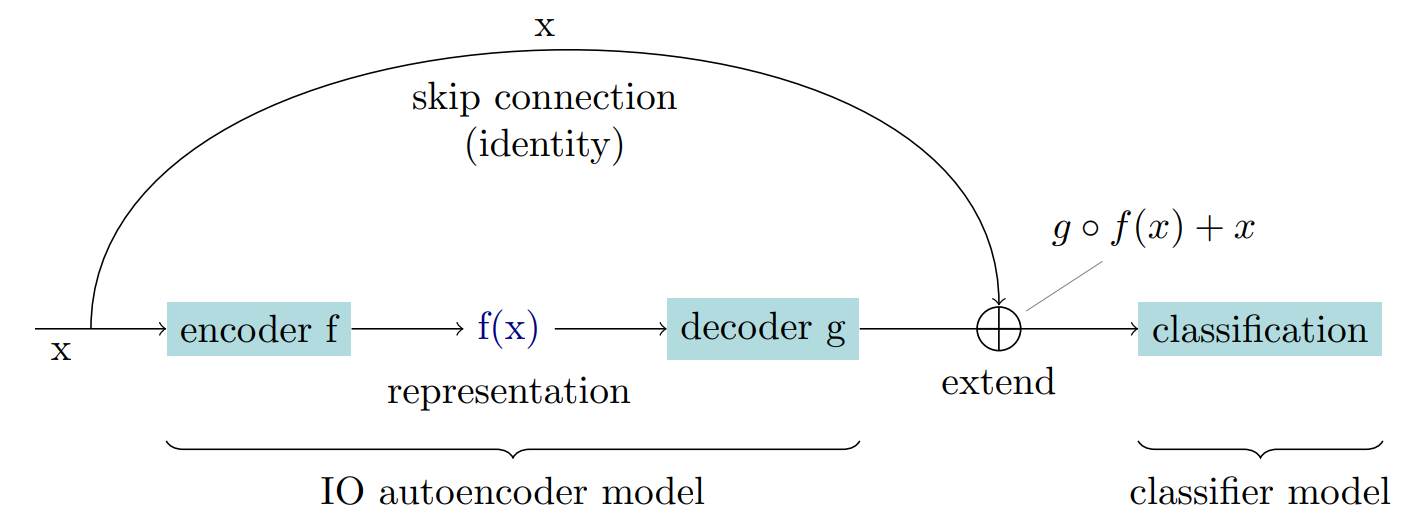}
        \caption{Models that appear in the MICS-EFS algorithm: IO autoencoder and classifier model.}
    \label{fig:model}
\end{figure*}

\noindent feature selection and input-output configuration search contributes to the overall effectiveness and efficiency of the MICS-EFS algorithm.

\vspace{0.5cm}
\noindent\textbf{Embedded feature selection.}
The objective of feature selection is to obtain a subset $S \subseteq \{1, 2, ..., n\}$ of features with a specified size $k$, and learn a reconstruction function $h: \mathbb{R}^k \mapsto \mathbb{R}^n$. This function minimizes the expected loss between the original sample x and the reconstructed sample $h(x_S)$, where $x_S \in R^k$ consists of those elements $x_i$ such that $i \in S$. Formally, this can be described as

\begin{equation}\label{eq3}
argmin_{h, S}E[L(x, h(x_s))].
\end{equation}

\noindent In general, selecting the optimal feature subset is a computationally challenging problem known as NP-hard \cite{amaldi1998approximability}.  While this complexity poses a significant challenge, the proposed MICS-EFS algorithm mitigates it through its integration with the SFS methodology. Specifically, instead of exhaustively evaluating all possible feature subsets, the algorithm employs a greedy, iterative strategy.

Drawing an analogy between feature selection and input-output configuration search, the optimization problem can be modified as 

\begin{equation}\label{eq4}
argmin_{h,x_{in}}E[L(x_{out}, h(x_{in}))],
\end{equation}

\noindent where $x_{in}$ represents the input and $x_{out}$ represents the output feature space of the original data. Equation \ref{eq2} and \ref{eq4} exhibit similarities that trace back to the same underlying problem when $h = g\circ f$. This similarity suggests that the input configuration search algorithm can serve as an alternative approach to feature selection algorithms, as both methods aim to identify informative and descriptive features or inputs.

\subsubsection{Modeling methodology: convolutional neural network}
\label{sec:modeling}
In the MICS-EFS algorithm, two models are employed. One model is responsible for exploring input-output configuration, as discussed in Section \ref{sec:input-output}. This model can also be referred to as the IO autoencoder model, whose objective is to optimize the model for any user-defined input-output configuration.  The second model is the classifier model, which employs a convolutional neural network (CNN). This section provides an in-depth explanation of the CNN-based classifier model.

\vspace{0.2cm}
\noindent\textbf{Classifier model.} The MICS-EFS modeling methodology incorporates a component based on a CNN. The choice of CNN is primarily motivated by its wide adoption, superior parameter estimation capabilities, and its ability to process various types of data, including 1D, 2D, and multimodal data. Moreover, the utilization of a CNN aligns with the IO autoencoder model, which is also CNN-based. For this research, a simple and shallow CNN architecture is utilized, emphasizing simplicity and transparency. It is important to note that this choice can be substituted with any other learning algorithm or even deeper and more complex CNN architectures. The goal of this CNN-based component is to perform task-specific classification, which also serves as the termination criterion for the SFS search algorithm discussed in Section \ref{sec:input-output2}. The search algorithm continues as long as improvements are achieved, and it stops when no accuracy improvement is observed, aiming to determine the optimal number of input features, if such exists.

\vspace{0.2cm}
\noindent\textbf{Novel evaluation system.} 
A novel evaluation system is introduced in the MICS-EFS algorithm that enhances the classification accuracy by integrating reconstructed data from the IO autoencoder into the evaluation process. Unlike traditional evaluation methods that solely rely on raw input data, this approach combines the original input with its reconstruction, thereby leveraging the autoencoder’s ability to capture internal dependencies and reduce noise. This modification enables the utilization of internal information within the input data. The algorithm proves to be particularly effective also in scenarios where the data contains significant noise or redundant information. With this modification, the algorithm becomes more robust. It is crucial to note that during the evaluation process, the input (x) remains unchanged, furthermore, the nature of the task, i.e., classification, remains the same (Figure \ref{fig:model}). Components are: 

\begin{itemize}
  \item[--] \textit{IO Autoencoder:} Reconstructs the data using the selected features, highlighting their relevance and the relationships among features.
\vspace{0.15cm}
  \item[--] \textit{Classifier:} Both the original input and the reconstructed data are simultaneously fed into the CNN-based classifier
\vspace{0.15cm}
  \item[--] \textit{Feature Validation:} The evaluation system validates the effectiveness of selected features by measuring their contribution to the classification accuracy.
\end{itemize}

\subsection{The novel algorithm of MICS-EFS}
\label{sec:idenasalg}

In this section, the MICS-EFS algorithm will be explained, highlighting and summarizing its fundamental characteristics and procedures. The input to the algorithm is a dataset D containing potential features, either it can be 1D signal or 2D image data, while the output is the optimal input-output configuration and the corresponding selected feature subset S that maximizes classification accuracy. The MICS-EFS algorithm employs a sequential forward search methodology as its primary mechanism. This iterative approach considers multiple  configurations and their associated relationships in each iteration. In one iteration, the algorithm aims to determine the optimal configuration that minimizes the reconstruction cost. Once the optimal configuration is identified, the algorithm proceeds to expand the input feature set while conducting evaluation using the classifier model. The search algorithm continues until a point is reached where further expansion does not yield significant improvements. The detailed steps of the proposed algorithm can be seen in Algorithm \ref{alg:idenas}.

\begin{algorithm}[H]
\hspace*{2.5mm} \textbf{Input:} dataset D, IO autoencoder AE, classifier C

\hspace*{2.5mm} \textbf{Output:} relevant feature set S
\begin{algorithmic}[1]

\State Initialize the input feature set S
\Repeat
    \For{each possible input feature in D}
        \State Evaluate the reconstruction cost using the AE
    \EndFor
    \State Select the input feature with the minimum cost
    \State Expand the input feature set S accordingly
    \State Perform evaluation using the C
\Until{no further improvement in accuracy or $|S| == k$}
\State Return the final input feature set S

\end{algorithmic}
\caption{Algorithm of the Input-Output configuration search}\label{alg:idenas}
\end{algorithm}

\noindent The algorithm offers a systematic approach to enhance the performance of the algorithm and achieve effective classification estimations. The source code of the software used in the research is accessible on GitHub: \url{https://github.com/viharoszsolt/IDENAS}

\subsection{Computational Complexity}

The computational complexity of the MICS-EFS algorithm is determined by its core components: the SFS, the autoencoder-based reconstruction, and the classifier evaluation. Each of these contributes to the overall complexity as follows:

\begin{itemize}
  \item[--] \textit{Sequential Forward Search:} The SFS methodology iteratively adds features to the selected subset while evaluating the reconstruction loss for each candidate feature. Given $n$ features, the process requires evaluating the remaining features at each iteration. In the worst case, this results in a complexity of $O(n^2)$ for the SFS component.
  \item[--] \textit{Autoencoder Reconstruction:} In each iteration, the reconstruction loss is calculated using an autoencoder. The complexity of this step depends on the dataset size $N$ (number of samples) and the number of parameters $P$ in the autoencoder. The forward pass through the network has a complexity of $O(N \cdot P)$ per iteration.
  \item[--] \textit{Classifier Evaluation:} The selected feature subset is validated using e.g. a CNN classifier. The evaluation complexity depends on the dataset size $N$ and the number of parameters $Q$ in the CNN. This step adds $O(N \cdot Q)$ to the computational cost per iteration.
\end{itemize}

\noindent Considering the interaction between these components, the overall computational complexity of the MICS-EFS algorithm can be expressed as:
\begin{equation}
O(n^2 \cdot (N \cdot P + N \cdot Q)) = O(n^2 \cdot N \cdot (P + Q)).
\end{equation}

In this expression, $n$ represents the number of features, $N$ the number of samples, $P$ the autoencoder parameters, and $Q$ the classifier parameters. The SFS component $O(n^2)$ dominates due to the iterative nature of the feature evaluation process.

\vspace{0.2cm}
\noindent\textit{Practical Considerations}: 
In practice, $n$ and $N$ are the primary drivers of complexity, as $P$ and $Q$ are typically fixed based on the architecture of the autoencoder and classifier. Additionally, calculation parallelization techniques may be employed to reduce runtime by evaluating multiple features or configurations simultaneously, although this does not affect the theoretical complexity.

\section{Experimental results and analysis}
\label{sec:results}

\subsection{Validation datasets}
\label{sec:data}
In this study, a comprehensive experimental evaluation was conducted on a set of benchmark classification datasets, encompassing both image data and sensor time-series data, with the majority of the signals related to human activity recognition tasks. A total of 7 image datasets and 6 sensor time-series datasets were utilized for this purpose. The datasets exhibited variations in terms of the number of instances, ranging from 120 to 70,000, the number of features, ranging from 24 to 3,072, and the number of classes, ranging from 5 to 17. Table \ref{dataset} provides an overview of the key details of each dataset. Additionally, the proposed method was validated in a real-world industrial application involving machining processes. This application demonstrated the practicality and effectiveness of the method in solving complex input-output configuration challenges in an industrial setting, further supporting its robustness and adaptability to real-world scenarios.

\begin{table}[h]
\caption{Description of the experimental datasets}\label{dataset}
\begin{tabular*}{250pt}{@{} cccccc@{} }
\toprule
\# & &Dataset & Instances & Features & Classes\\
\midrule
1 & \parbox[t]{2mm}{\multirow{4}{*}{\rotatebox[origin=c]{90}{image}}} &MNIST & 70000 & 784 & 10\\
2 & &MNIST (with noises) & 70000 & 784 & 10 \\
3 & &Fashion MNIST & 70000 & 784  & 10\\
4 & &CIFAR10 & 60000 & 3072 & 10 \\
5 & \parbox[t]{2mm}{\multirow{6}{*}{\rotatebox[origin=c]{90}{signal}}}&MHealth & 120 & 23 & 12 \\
6 & &HAR Smartphone & 10299 & 561 & 6 \\
7 & &Opportunity & 2551 & 242 & 5 \\
8 & &Skoda Mini Checkpoint & 700 & 60 & 10 \\
9 & &MIT-BIH & 48 & 30 & 5 \\
10 & &Measured cutting (industrial) & 120 & 7 & 10 \\
\bottomrule
\end{tabular*}
\end{table}

\subsection{Data pre-processing}
\label{sec:preprocess}

\begin{figure*}[h!]
    \centering
        \includegraphics[width=0.8\linewidth]{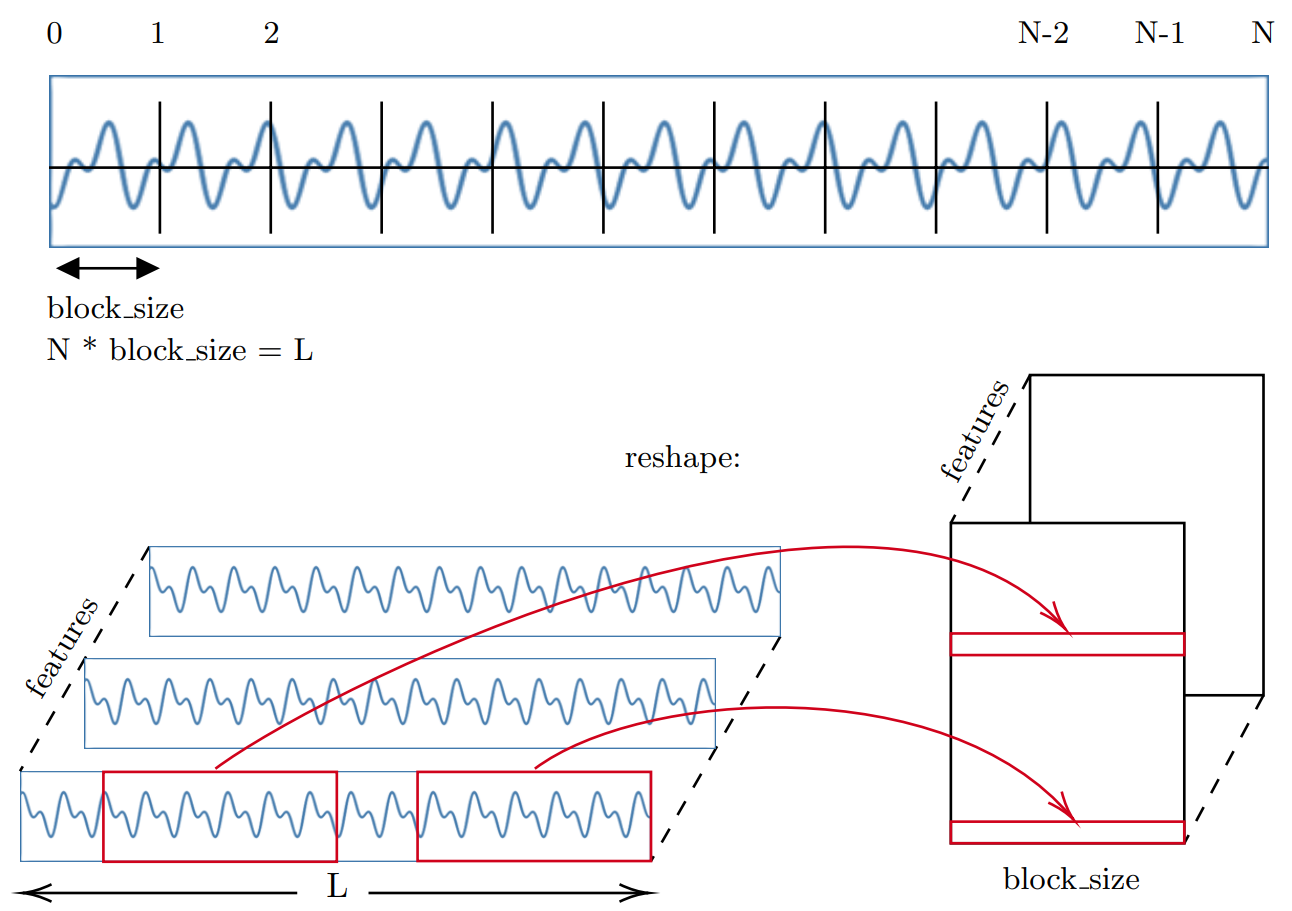}
        \caption{Sensor time-series data pre-processing for the MICS-EFS algorithm.}
    \label{fig:preprocessing}
\end{figure*}

The image and signal datasets underwent pre-processing steps prior to the utilization of the MICS-EFS model. For image data, a relatively straightforward pre-processing procedure was employed. The data underwent normalization: min-max normalization was applied to 2D image data, while standard normalization proved effective for 1D sensor time-series data. In the case of sensor time-series data, as each channel of the sensors has a different type of nature, the data was normalized not entirely, but separately for each individual channel. When dealing with time-series data,  a window slicing strategy was implemented to divide the data into smaller chunks (Figure \ref{fig:preprocessing}). This approach served two objectives: improving the training performance by working with shorter time-series data and increasing the number of instances, thereby augmenting the dataset size for training purposes. In some datasets, certain data cleaning and elimination techniques were applied as part of the pre-processing stage to ensure data quality. Finally, the labels were transformed into a one-hot encoded representation, enhancing the expressiveness and interpretability of the categorical data.

\subsection{Experimental setup}
\label{sec:settings}
\noindent\textbf{Compared algorithms.} 
A comparative analysis was conducted to evaluate the performance of the proposed MICS-EFS algorithm demonstrating its potential superiority. The algorithm was compared with seven existing unsupervised feature selection algorithms, already mentioned in Section \ref{sec:related}, namely: AEFS \cite{han2018autoencoder}, AgnoS \cite{doquet2020agnostic}, CAE \cite{balin2019concrete}, FAE \cite{wu2021fractal}, QS \cite{atashgahi2022quick}, UDRN \cite{zang2023udrn}.

\vspace{0.2cm}
\noindent\textbf{Evalution metrics.} 
During the research experiments, the performance evaluation of the proposed MICS-EFS algorithm encompassed two key aspects: accuracy and reconstruction error. The reconstruction error served as a measure of the algorithm's ability to accurately reconstruct the original input, thus, the relationships input-output configuration can be discovered. On the other hand, accuracy was employed as a metric to evaluate the algorithm's classification performance. In this study, only these metrics are presented. These metrics have been selected as they are widely recognized and commonly utilized in related studies for numerical comparisons.

\begin{figure}[H]
     \centering
         \includegraphics[width=0.8\linewidth]{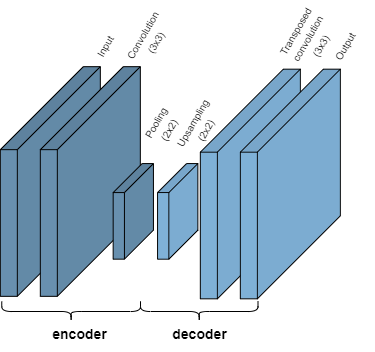}
         \caption{2D CNN Architecture of the applied IO autoencoder} 
         \label{fig:io-arch}
\end{figure}

\vspace{0.2cm}
\noindent\textbf{Experimental settings.}
In Section \ref{sec:modeling}, it was explained that the MICS-EFS algorithm consists of two models: 1) for input-output configuration search (IO autoencoder), and 2) for classification (classifier). The IO autoencoder used a simplified architecture with convolution, max pooling, upsampling, and transposed convolution layers. Kernel sizes were 3x3 or 4, determined by image and signal data, respectively, with pooling and upsampling layers at a factor of 2 (Figure \ref{fig:io-arch}). ReLU activation, RMSE cost function, and ADAM optimization \cite{kingma2014adam} with early stopping was employed.

\noindent The classifier model employed similarly, however, as it served a different task, the cross-entropy function was utilized as the cost function at the end, instead of having a decoder structure. Due to the relatively simple parameters of both models, there is the potential to explore in further research more complex architectures. However, for the purposes of the actual research, a simpler model was chosen to establish a baseline and evaluate its performance. For both models, the datasets were divided into training, validation, and test sets in a ratio of 67:22:11 unless predetermined. For testing the robustness aspect, the training process was repeated 30 times, and the results were statistically evaluated to calculate the reconstruction errors and classification accuracy. The intensive training was conducted using NVIDIA GeForce RTX 2080 and NVIDIA Tesla V100 graphics cards provided by the HUN-REN Cloud and the internal institute server (see Section \ref{sec:acknowledgement}).

\setlength{\intextsep}{10pt plus 2pt minus 2pt}

\subsection{Case study}
\label{sec:evaluationsteps}
This section provides a detailed information of the evaluation procedure's individual steps conducted on the MNIST dataset as example. Similar approach was adopted for evaluating all the datasets in the research presented in Table \ref{dataset}.

\begin{figure*}[h]
     \centering
         \includegraphics[width=10.5cm]{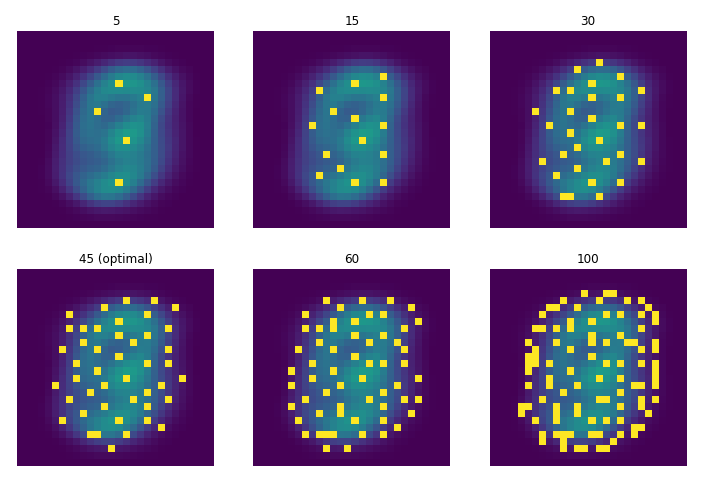}
         \caption{Results of the input search for the MNIST dataset for different number of features. Yellow color indicates the selected feature. The background shows the total averaged intensity of the MNIST dataset.} 
         \label{fig:features}
\end{figure*}

\vspace{0.2cm}
\noindent\textbf{Selected features.} The primary objective of the input-output configuration search is to identify and select the most informative input features within the given dataset, whether they are individual pixels in an image or specific sensors in signal data. These selected features possess a high level of descriptiveness, as they have the ability to accurately estimate the remaining features of the given pattern with minimal error. By focusing on these highly informative features, the subsequent stages of analysis and modeling can benefit from a more concise and representative set of inputs, contributing to enhanced overall performance and accuracy.
\noindent In Figure \ref{fig:features}, the most characteristic input pixels of the MNIST dataset are depicted for different number of features. It can be observed that the features are sparsely scattered, validating the independence of the selected ones. With an increase in the number of features, it is evident that the new features are situated more in the background, justifying the determination of an optimal number of features.

\begin{figure}[H]
     \centering
         \includegraphics[width=0.8\linewidth]{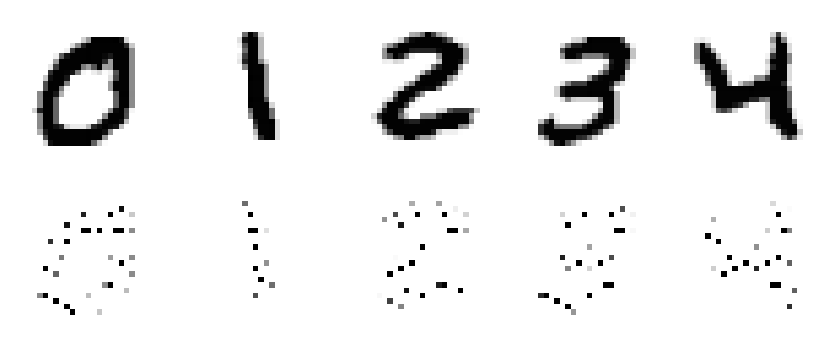}
         \caption{The image shows five examples from the MNIST dataset in the first row, and their characteristic (45) pixels, determined by MICS-EFS, in the second row.} 
         \label{fig:selected}
\end{figure}

\begin{figure}[H]
     \centering
         \includegraphics[width=0.5\textwidth]{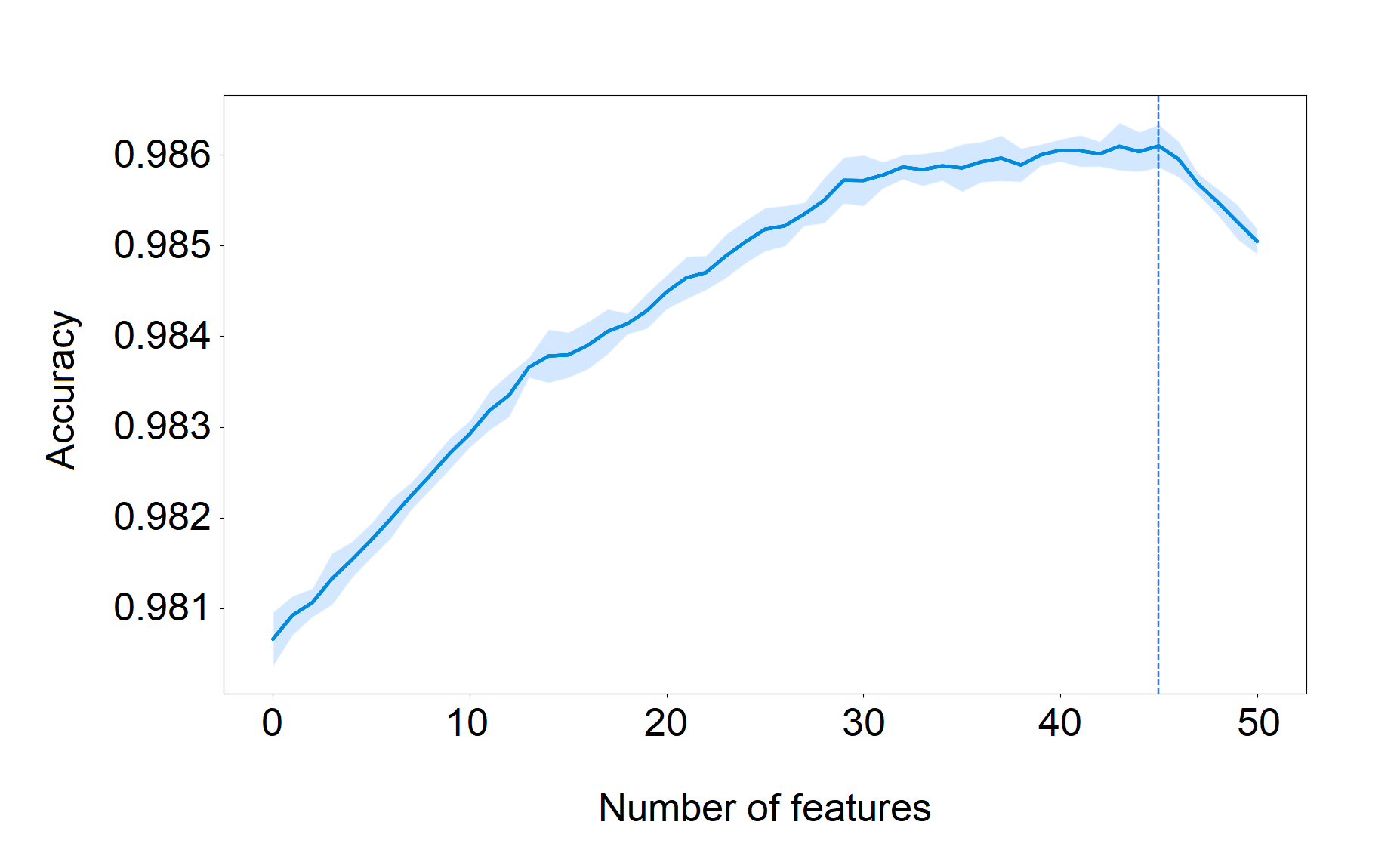}
         \caption{Determining optimal number of features for MNIST dataset. The standard deviation of the 30 classifications is marked with a pale blue color.} 
         \label{fig:features-opt}
\end{figure}

\noindent\textbf{Optimal number of features.} To identify the optimal number of features, a task-specific classification task was conducted as part of the evaluation process. Once a configuration was determined, a series of 30 classifications were performed, and the average accuracy was calculated. This iterative procedure was then repeated, gradually incorporating additional features into the analysis. The evaluation process continued until the point where the accuracy ceased to show further improvement beyond the best achieved result, ensuring that the selection of features reached an optimal and stable configuration.  As an example, Figure \ref{fig:features-opt} and \ref{fig:selected} show the result of such a procedure for the MNIST dataset, demonstrating that approximately 5\% of the features, specifically 45 features, are sufficient for effective classification.

\subsection{Comparison of baseline CNN}
\label{sec:superiority}
\begin{figure*}[h]
    \centering
    {\includegraphics[width=.3\textwidth]{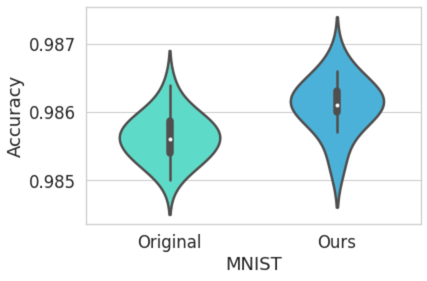}} \quad
    {\includegraphics[width=.3\textwidth]{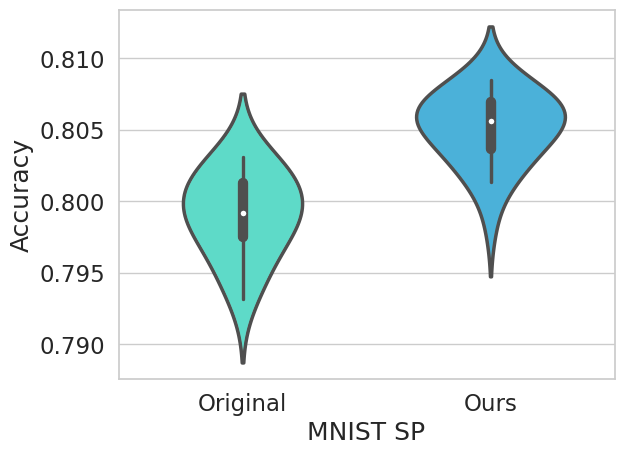}} \quad
    {\includegraphics[width=.3\textwidth]{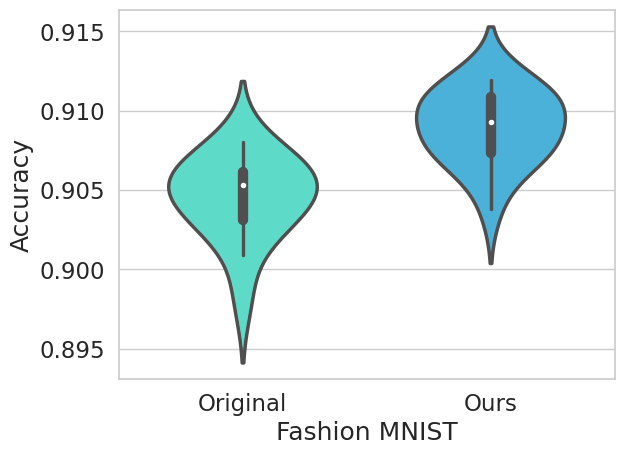}} \\
    {\includegraphics[width=.3\textwidth]{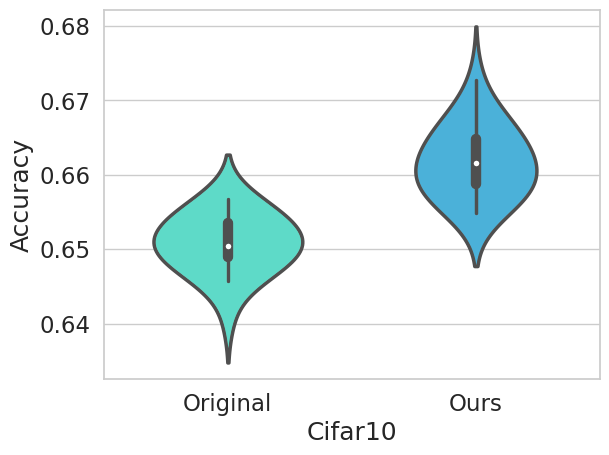}} \quad
    {\includegraphics[width=.3\textwidth]{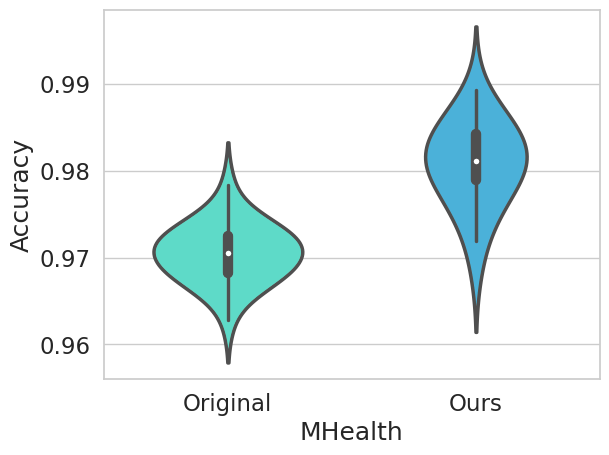}} \quad
    {\includegraphics[width=.3\textwidth]{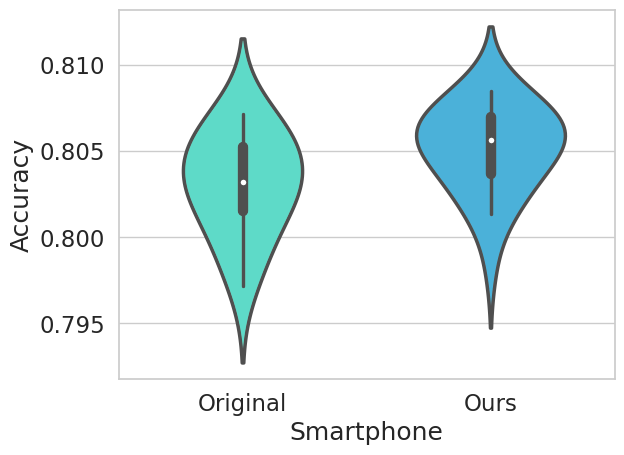}} \\
    {\includegraphics[width=.3\textwidth]{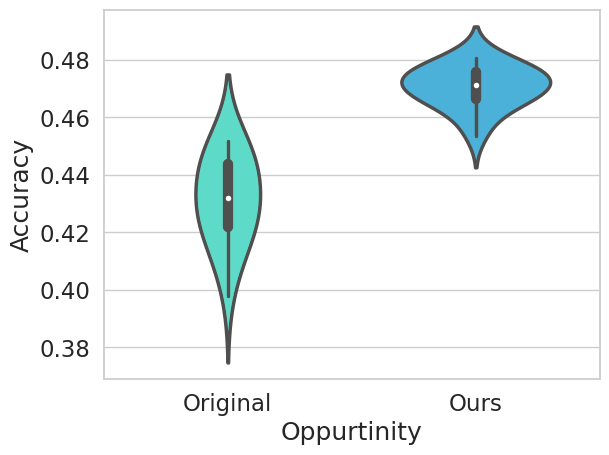}} \quad
    {\includegraphics[width=.3\textwidth]{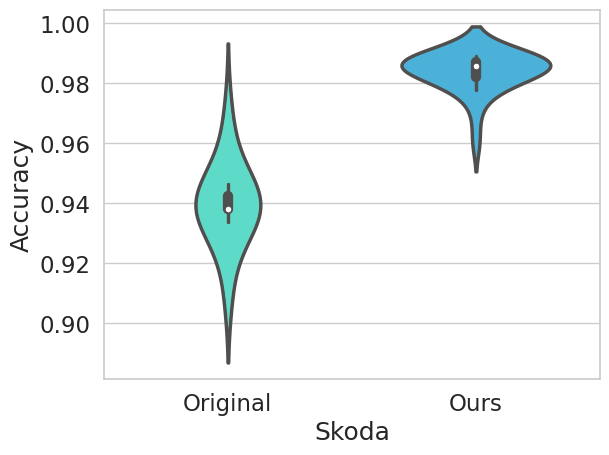}} \quad
    {\includegraphics[width=.3\textwidth]{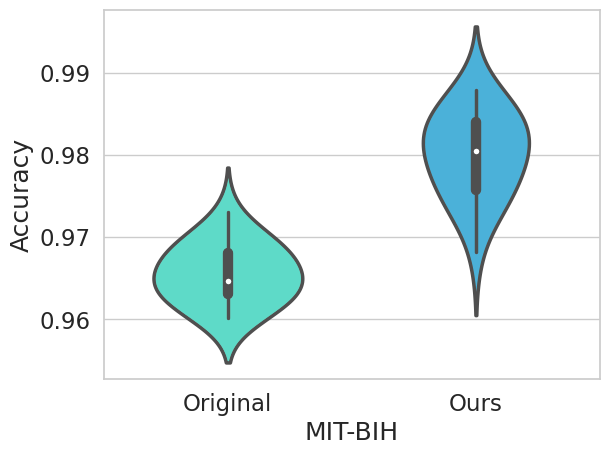}}
    \caption{Overall results of the proposed MICS-EFS algorithm. In the figure, a comparative analysis was performed for all the datasets selected and visualized in the form of violin plots. In any case, improvements in terms of classification accuracy have been achieved.}
    \label{fig:results}
    \end{figure*}

The algorithm's classification capability was assessed by comparing the accuracy against a shallow CNN with a very few convolutional and pooling layer, serving as a baseline.  The same model was applied in both approaches to ensure comparability. This quantitative evaluation was performed by conducting 30 independent runs of each classification task, and the average results were obtained (Table \ref{tab:acc2}). To facilitate a clearer understanding, the outcomes were visually represented using violin diagrams. Notably, for each dataset, the algorithm consistently outperformed the baseline, as demonstrated in Figure \ref{fig:results}.

The results demonstrate consistent improvements, with MICS-EFS achieving an average enhancement of 1.5\% across diverse classification tasks, including 0.5\% for image data and 2.57\% for sensor data, highlighting its superiority over state-of-the-art methods. Additionally, the algorithm requires only 2–5\% of the original feature set to deliver comparable or superior performance, substantially reducing computational requirements.

\vspace{0.2cm}
\begin{table}[h]
\centering
\caption{Classification accuracy (\%) for comparing our MICS-EFS method with shallow CNN network}\label{tab:acc2}
\begin{tabular*}{210pt}{@{} ccccc@{} }
\toprule
\# & Dataset & Reference CNN & MICS-EFS \\
\midrule
1 & MNIST & $98.5$ & $\pmb{98.6}$\\
2 & MNIST (with noise) & $79.8$ & $\pmb{80.5}$ \\
3 & Fashion MNIST & $90.5$ & $\pmb{90.9}$  \\
4 & CIFAR10 & $65.0$ & $\pmb{66.1}$ \\
5 & MHealth & $97.0$ & $\pmb{98.1}$\\
6 & HAR Smartphone & $80.3$ & $\pmb{80.6}$ \\
7 & Opportunity & $43.0 $& $\pmb{47.5}$ \\
8 & Skoda Mini Checkpoint & $93.8$ & $\pmb{98.9}$\\
9 & MIT-BIH & $96.3$ & $\pmb{98.0}$ \\
\bottomrule
\end{tabular*}
\end{table}

\vspace{0.2cm}

\begin{table}[h]
\centering
\caption{Classification accuracy (\%) with selected features by different algorithms}\label{tab:acc}
\begin{tabular*}{175pt}{@{} ccccc@{} }
\toprule
\# & Algorithm & MNIST & Fashion MNIST \\
\midrule
1 & AEFS & $80.2\pm2.6$ & $79.4\pm1.5$ \\
2 & UDFS & $88.1\pm1.6$ & $79.3\pm1.2$ \\
3 & PFA & $88.5\pm2.0$ & $80.3\pm10.5$  \\
4 & AgnoS-S & $43.5\pm15.6$ & $78.4\pm1.3$ \\
5 & CAE & $92.5\pm0.4$ & $82.3\pm1.0$ \\
6 & FAE & $92.9\pm0.7$ & $82.5\pm0.6$ \\
7 & QS & $93.2\pm0.2$ & N/A \\
8 & UDRN & $94.2\pm0.3$ & N/A \\
9 & MICS-EFS & $\pmb{98.6\pm3.2}$ & $\pmb{90.9\pm2.0}$ \\
\bottomrule
\end{tabular*}
\end{table}
\vspace{0.2cm}

\begin{figure*}[h]
     \centering
         \includegraphics[width=13cm]{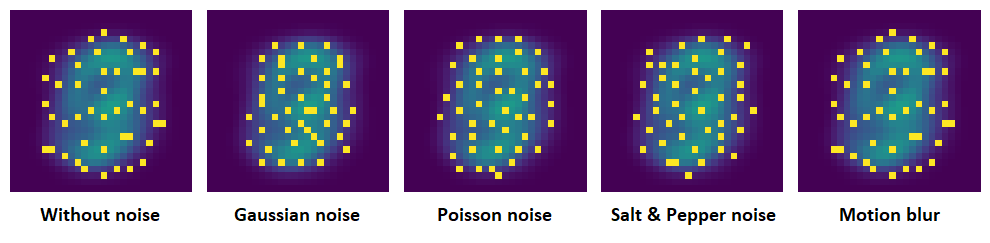}
         \caption{The image illustrates a total averaged MNIST dataset in the background, showing the selected features under different noise conditions, marked as yellow.} 
         \label{fig:noise2}
\end{figure*}

\begin{figure*}[h!]
     \centering
         \includegraphics[width=13cm]{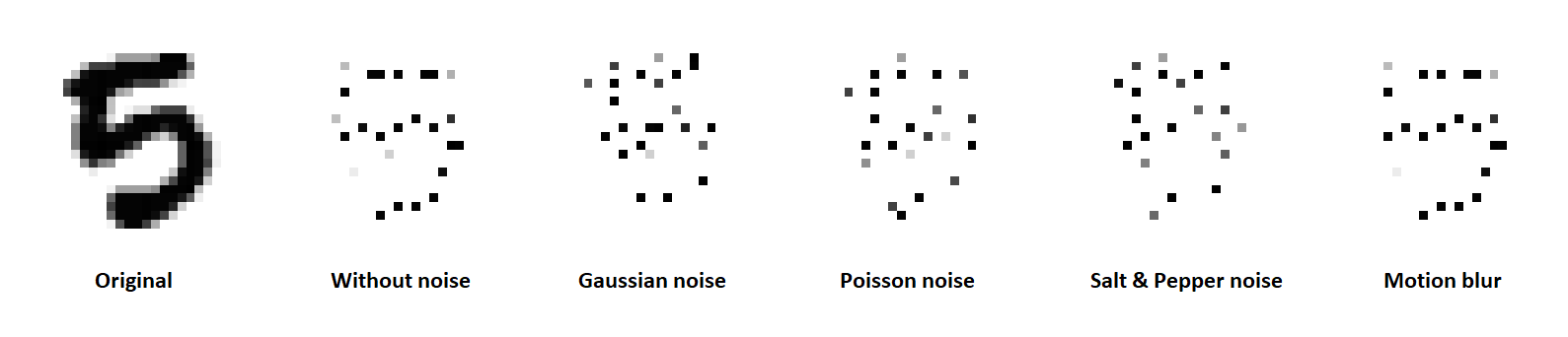}
         \caption{The image illustrates an original MNIST digit and its variations, showing the selected features under different noise conditions.} 
         \label{fig:noise}
\end{figure*}

\newpage
\subsection{Comparison of FS methods}
\label{sec:superiority2}
The second approach involved comparing the algorithm's performance against existing autoencoder-based feature selection algorithms, as mentioned in Section \ref{sec:settings}. This analysis focused on evaluating the accuracy. Due to the limited availability of shared datasets among these algorithms, the evaluation was conducted on a few datasets, namely MNIST and Fashion MNIST. Those two datasets are the most frequently used for comparison. The results of this analysis are presented in Table \ref{tab:acc}, providing a comprehensive overview of the algorithm's performance in comparison to the existing methods. Through these quantitative evaluations and comprehensive analyses, the MICS-EFS results underscore its superiority over baseline approaches in terms of classification accuracy, while also showcasing its competitive performance when compared to established autoencoder-based feature selection algorithms.


\subsection{Algorithm's robustness}
\label{sec:robustness}
The robustness of the proposed MICS-EFS algorithm was evaluated using the MNIST dataset under various noise conditions, including Gaussian noise, Poisson noise, Salt \& Pepper noise, and Motion Blur \cite{mu2019mnist}. The primary objective of this analysis was to assess the algorithm's ability to maintain performance in the presence of corrupted data. 

\noindent Quantitative Results: While Figure \ref{fig:noise2} and \ref{fig:noise} provide a qualitative visualization of the selected features under different noise conditions, quantitative metrics such as classification accuracy, reconstruction error, and feature overlap percentage can offer deeper insights into the algorithm's robustness.

\begin{itemize}
  \item[--] \textit{Classification Accuracy:} The MICS-EFS algorithm demonstrates minimal degradation in accuracy across most noise conditions, with variations remaining within $\pm5\%$ of the baseline performance. However, under Salt and Pepper noise, a more significant reduction in accuracy was observed, highlighting the algorithm's sensitivity to this specific type of noise. Despite this, the algorithm maintains robust performance in the presence of other noise types, reflecting its general adaptability to challenging environments.
  \item[--] \textit{Reconstruction Error:} RMSE increase moderately with higher noise levels but remain significantly lower than those observed with other FS methods, highlighting the algorithm’s ability to focus on relevant features. 
  \item[--] \textit{Feature Overlap:} The percentage of overlapping features selected under noisy and noise-free conditions averages 45\%. While this overlap may not be exceptionally high, the algorithm's ability to achieve strong accuracy and RMSE results despite this indicates that the feature selection process effectively identifies the most relevant features, even in the presence of noise.
\end{itemize}

These results suggest that MICS-EFS is robust to various noise types, effectively identifying and utilizing the most descriptive features even under challenging conditions. The visualization in Figure \ref{fig:noise2} and \ref{fig:noise} complement these findings by illustrating that selected features align with regions of high information density, regardless of noise interference.

\subsection{Visualization of sensory data for Human Activity Recognition}
\label{sec:explainResult}
\begin{figure*}[h]
\centering
\begin{subfigure}{0.9\textwidth}
\centering
\includegraphics[width=1\linewidth]{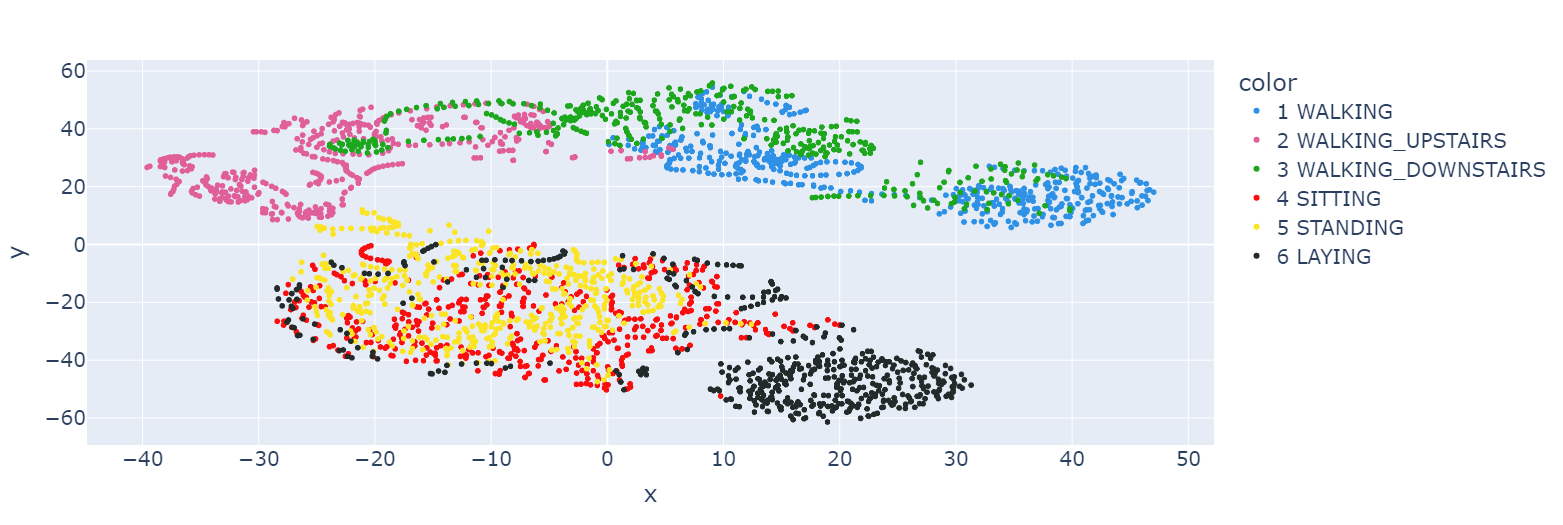}
\caption
{\centering 2D visualization of the classical convolutional neural network}
\label{fig:smart12}
\end{subfigure}

\begin{subfigure}{0.9\textwidth}
\centering
\includegraphics[width=1\linewidth]{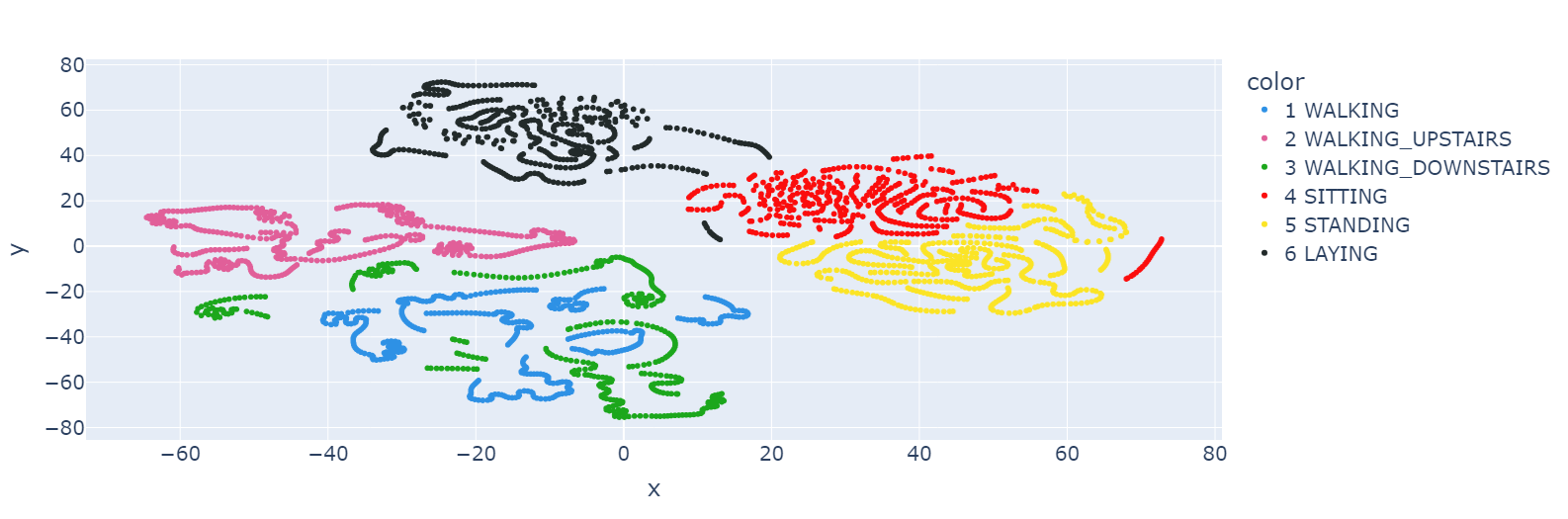}
\caption
{\centering 2D visualization of the proposed MICS-EFS method}
\label{fig:smart11}

\end{subfigure}

\caption
{2D visualization of T-SNE results on the Smartphone sensory dataset. The top image shows classification by a classical CNN, while the bottom depicts results from the proposed MICS-EFS method on the same dataset.}
\label{fig:smart}
\end{figure*}

For visualization purposes, the T-SNE (t-Distributed Stochastic Neighbor Embedding) dimensionality reduction procedure \cite{van2008visualizing}, as one of the most frequently applied techniques \cite{gisbrecht2015parametric}, was employed, enabling a 2D representation of the learned dependencies. By comparing the classification results of the shallow convolutional neural network (CNN) architecture with the proposed MICS-EFS solution, both trained on the same smartphone dataset, it was proved that the proposed algorithm outperforms in terms of producing a more distinct and clearer separation (Figure \ref{fig:smart}). This visual representation solidifies the credibility, effectiveness and superiority of the proposed algorithm for the smartphone dataset.

\subsection{Industrial Validation of the Proposed Method}
\label{industry}
The proposed MICS-EFS algorithm was validated using a dataset measured from cutting processes in an industrial environment, emphasizing its applicability to real-world scenarios. The dataset comprises parameters collected during machining operations, specifically focused on material removal processes such as turning. Seven parameters were included in the dataset: feed rate (f, mm/rev), depth of cut (a, mm), cutting speed (v, m/min), cutting force (Fc, N), cutting power (P, kW), temperature (T, °C), and surface roughness (Ra, mm).

A total of 120 samples were measured under controlled conditions, capturing variations across different machining setups. Data preprocessing included normalization, cleaning, and sliding window sampling to handle noise and anomalies. The dataset was particularly structured to enable the analysis of dependencies between input and output variables, highlighting the non-linear and multidimensional nature of the machining parameters. This dataset serves as a foundation to demonstrate the efficiency and accuracy of the proposed method in industrial applications.

\begin{figure*}[h!]
     \centering
         \includegraphics[width=0.8\linewidth]{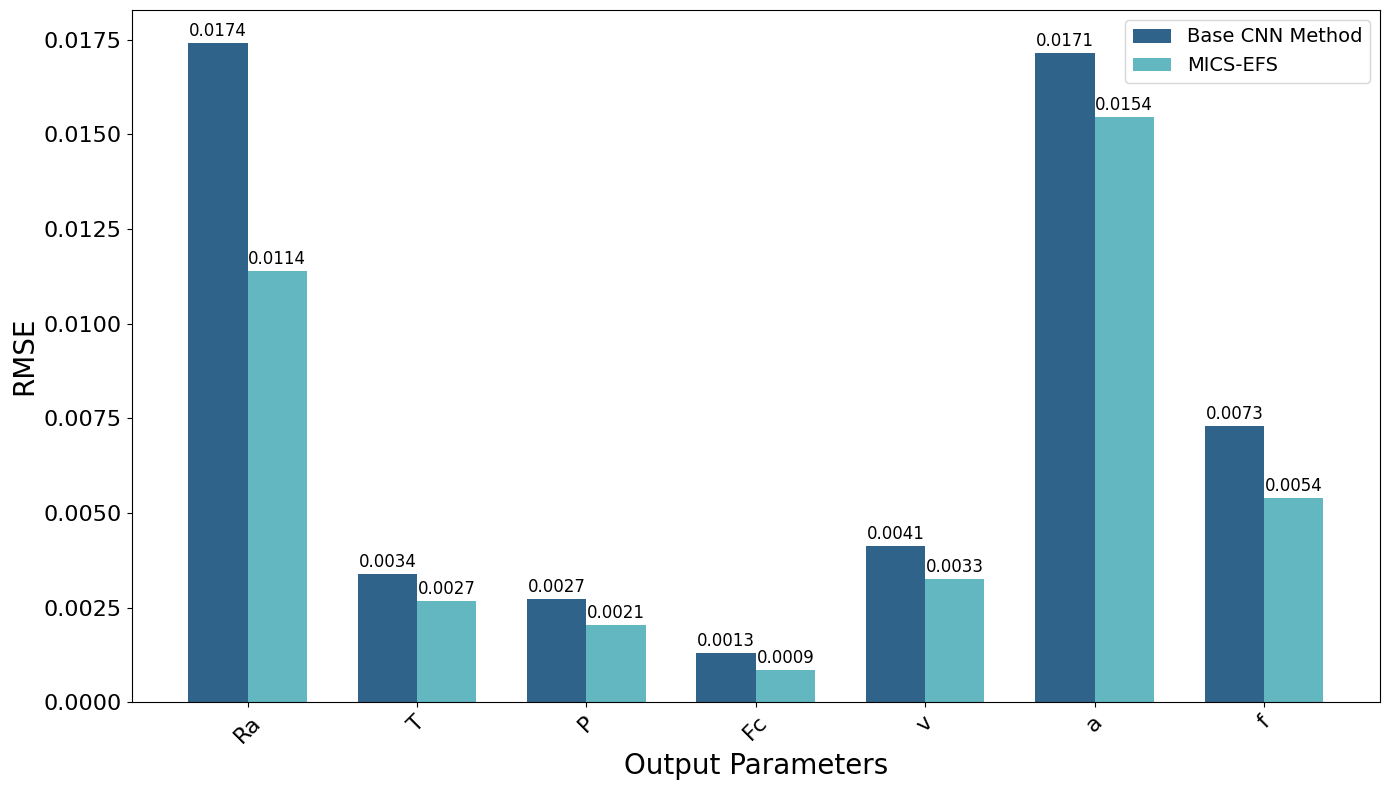}
         \caption{The bar chart compares the RMSE values of the reference CNN (in dark blue) and the Proposed MICS-EFS (in light teal) across various output parameters, including Ra, T, P, Fc, v, a, and f.} 
         \label{fig:inext}
\end{figure*}

A comparison of RMSE values for various output parameters between the traditional CNN model and the proposed MICS-EFS method is shown in Fig \ref{fig:inext}. Across all output parameter, the proposed MICS-EFS consistently achieves lower RMSE values, indicating superior performance in accurately identifying the input configurations and relationships between the output. This demonstrates the effectiveness of MICS-EFS in reducing errors and improving prediction accuracy. The results further validate the robustness of MICS-EFS, particularly in industrial and real-world applications, such as machining processes, where precise parameter optimization is critical for performance. These findings highlight the potential of MICS-EFS as a practical tool for solving complex input-output configuration challenges in diverse application domains.

\section{Conclusions and future works}
\label{sec:conclusion}

The paper introduced the novel Model Input-Output Configuration Search with Embedded Feature Selection (MICS-EFS) algorithm, which seamlessly integrates a novel feature exploration approach with Neural Architecture Search (NAS). The approach addresses the limitations of traditional NAS algorithms by autonomously determining optimal model input-output configurations, providing an improved alternative to scenarios where the input-output mapping are not readily available. 

The motivation behind the proposed approach arise from the evolving landscape of neural networks and the growing significance of neural architecture search. Traditional NAS algorithms focus solely on the internal structure, between the input and output layer, considering inputs and output as fixed parameters. Classical feature selection methods, while exploring the relevant input features do not modify the input-output structure of the models completely, especially, since they do not select variables among the output parameters. While some preliminary work exists for shallow neural networks, MICS-EFS introduces this concept to the deep learning domain. As a significant extension, the proposed MICS-EFS algorithm introduces a groundbreaking two-level approach, autonomously exploring dependencies among inputs to dynamically adapt the input-output structure.

The MICS-EFS methodology employs Sequential Forward Search (SFS) as the chosen search algorithm, offering simplicity and transparency in the exploration process. An embedded feature selection mechanism operates silently, identifying relevant features and providing rankings. MICS-EFS uses a specifically designed encoder-decoder model to explore input-output configuration, especially for 1D sersor time-series and 2D image data. 

A comprehensive experimental evaluation was conducted on a set of representative benchmarking classification datasets. A total of 7 image datasets and 6 signal datasets were utilized for this purpose. Additionally, the proposed method was validated in a real-world industrial application involving machining processes. This application demonstrated the practicality and effectiveness of the method in solving complex input-output configuration challenges in an industrial setting, further supporting its robustness and adaptability to real-world scenarios. During the research, both quantitative and qualitative analyses were conducted. \textbf{The results exhibit consistent improvements, with MICS-EFS achieving a notable average enhancement of 1.5\% across various classification tasks, 0.5\% and 2.57\% for image and sensor data, respectively, demonstrating its superiority over state-of-the-art solutions. Moreover, the method requires only 2–5\% of the original data to achieve comparable or better results, significantly reducing computational demands.}

The source code of the software developed in the research has been published to GitHub as well: \url{https://github.com/viharoszsolt/IDENAS}



\subsection{Future works}
\label{sec:future}
There are several, additional, potential directions for MICS-EFS that merit exploration. One avenue involves the exploration and application of alternative, more sophisticated, more complex configuration search model types. The adoption of such models holds the potential to provide more efficient solutions and further increase the additional accuracy improvement of MICS-EFS, resulting in the expansion of the applicability of embedded feature selection algorithm(s) to diverse datasets, e.g., for handling text-based data as well. This extension could significantly enhance the adaptability and performance of these algorithms across different domains. This direction of the research has been already started by the authors.

Another prospective area for future research lies in the utilization of alternative search algorithms that diverge from greedy strategies. Implementing non-greedy algorithms introduces the possibility of discovering alternative and potentially superior (also local, but improved) optimum. By employing algorithms with a broader exploration scope, researchers can uncover hidden patterns and relationships within the data that may be overlooked by more deterministic approaches.


\section*{Data Availability Statement}
\label{sec:dataav}
All the datasets analyzed during the current study can be downloaded using the following links: MINST dataset (\url{https://yann.lecun.com/exdb/mnist/});
Fashion MNIST (\url{https://github.com/zalandoresearch/fashion-mnist});
CIFAR10 (\url{https://www.cs.toronto.edu/~kriz/cifar.html});
MHEALTH (\url{https://archive.ics.uci.edu/dataset/319/mhealth+dataset});
Smartphone (\url{https://archive.ics.uci.edu/dataset/240/human+activity+recognition+using+smartphones});
Opportunity (\url{https://archive.ics.uci.edu/dataset/226/opportunity+activity+recognition});
Skoda (\url{http://har-dataset.org/doku.php?id=wiki:dataset});
MIT-BIH (\url{https://physionet.org/content/mitdb/1.0.0/});
The Measured Cutting datasets that support the findings of this study are available from the authors upon request.

The source code of the software developed in the research has been published to GitHub as well: \url{https://github.com/viharoszsolt/IDENAS}

\section*{Conflict of Interest}
\label{sec:coi}
All authors have no conflicts of interest.

\section*{Acknowledgment}
\label{sec:acknowledgement}
This research has been supported by the European Union project RRF-2.3.1-21-2022-00004 within the framework of the Artificial Intelligence National Laboratory and by the TKP2021-NKTA-01  NRDIO grant on "Research on cooperative production and logistics systems to support a competitive and sustainable economy".

On behalf of Project "Comprehensive Testing of Machine Learning Algorithms" we thank for the usage of HUN-REN Cloud (https://science-cloud.hu/) that significantly helped us achieving the results published in this paper.

\bibliographystyle{plain}
\bibliography{sn-bibliography}

\newpage
\section*{Appendix A. Selected features for image dataset}
\label{sec:appendixa}

\begin{minipage}{1\textwidth} 
\begin{figure}[H]
\centering
\includegraphics[width=14cm]{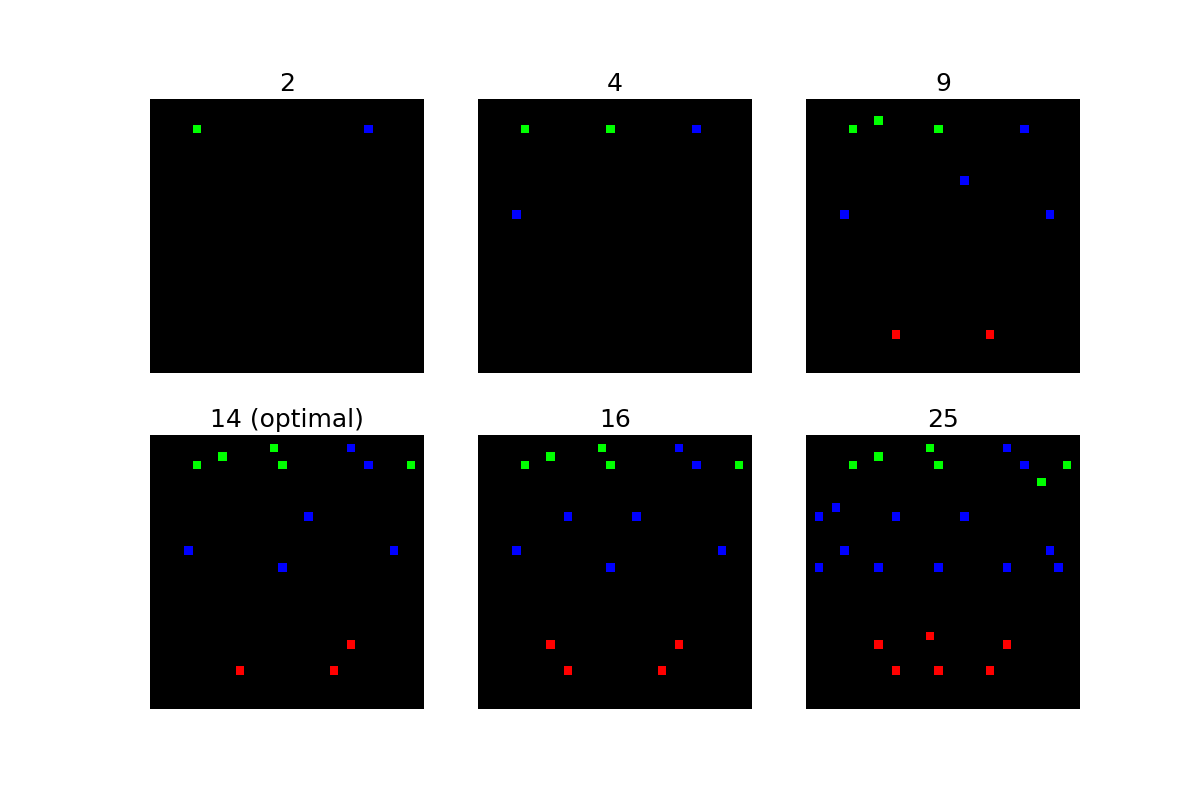}

\caption{\label{fig:label} Results of the input search for the CIFAR10 dataset for different number of features. The selected features from different channels are marked with distinct colors.}
\end{figure}

\begin{figure}[H]
\centering
\includegraphics[width=14cm]{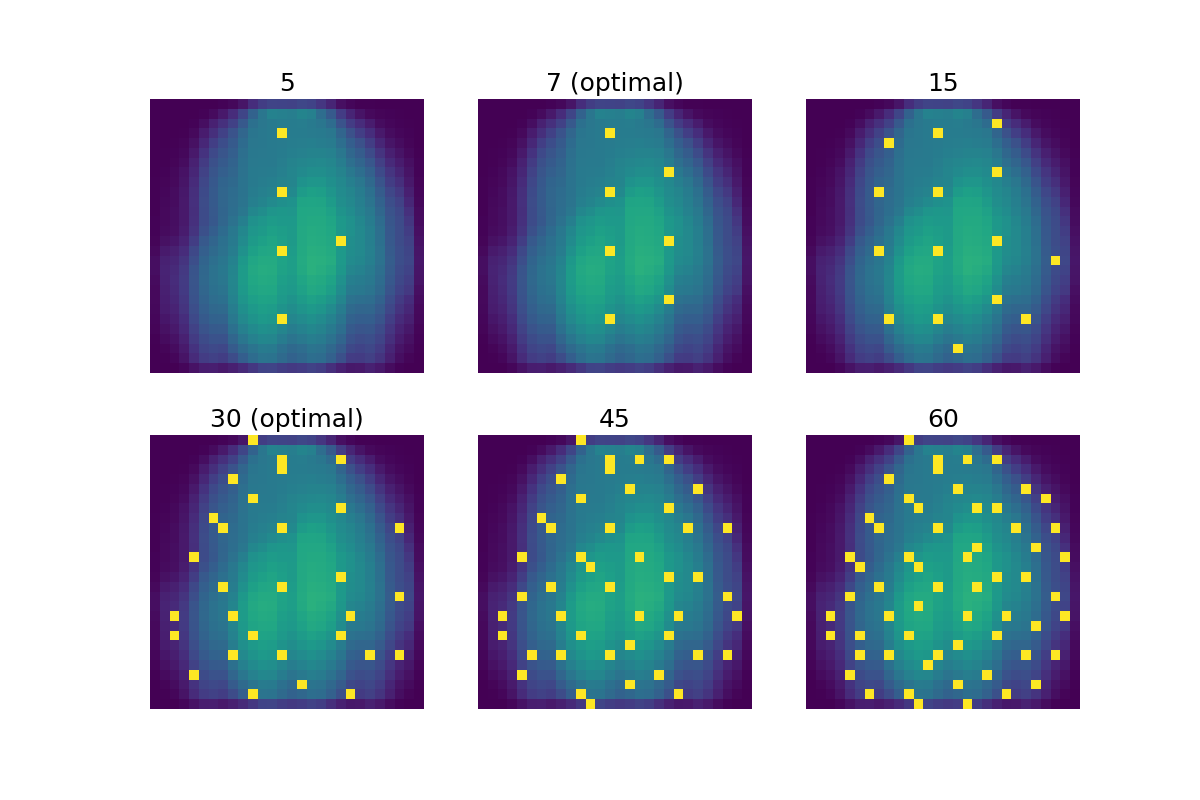}

\caption{\label{fig:label} Results of the input search for the Fashion MNIST dataset for different number of features. Yellow color indicates the selected feature/pixel. A case of 7 features is optimal according to our modeling methodology. The background shows the averaged intensity of the MNIST Fashion dataset.}
\end{figure}

\end{minipage}

\newpage
\clearpage
\newpage
\section*{Appendix B. 2D visualization for signal dataset}
\label{sec:appendixb}

\begin{minipage}{1\textwidth} 
\begin{figure}[H]
\centering
\includegraphics[width=16cm]{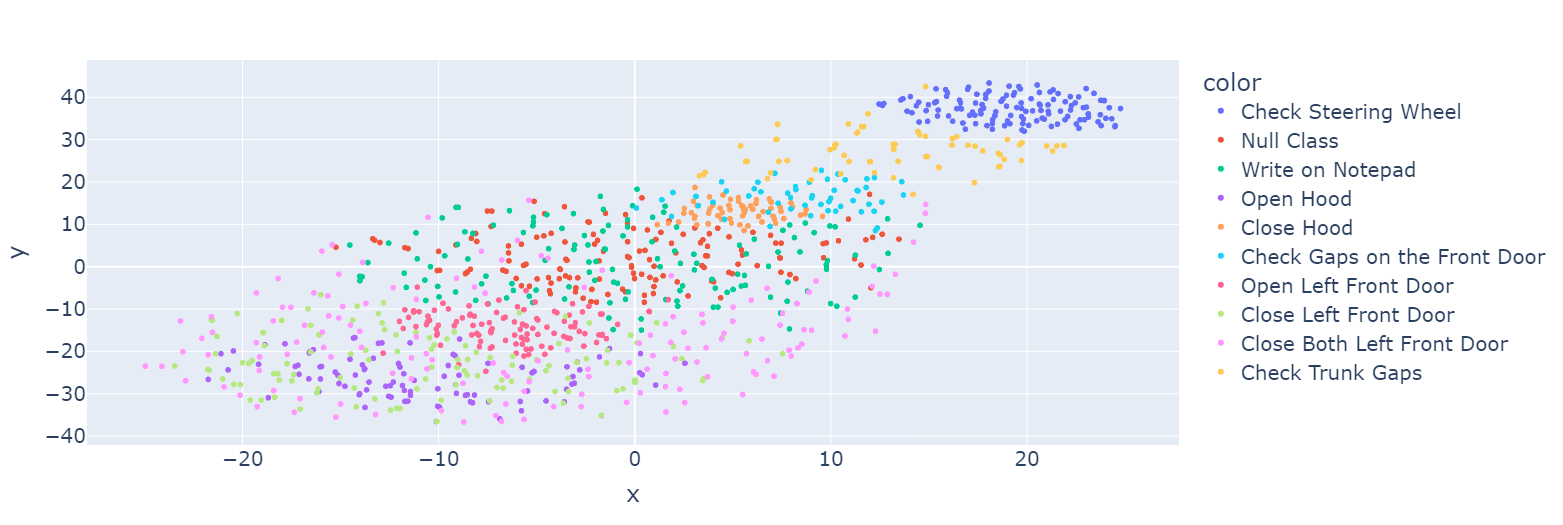}

\subcaption{\label{fig:label} 2D visualization of the classical convolutional neural network}
\end{figure}

\begin{figure}[H]
\centering
\includegraphics[width=16cm]{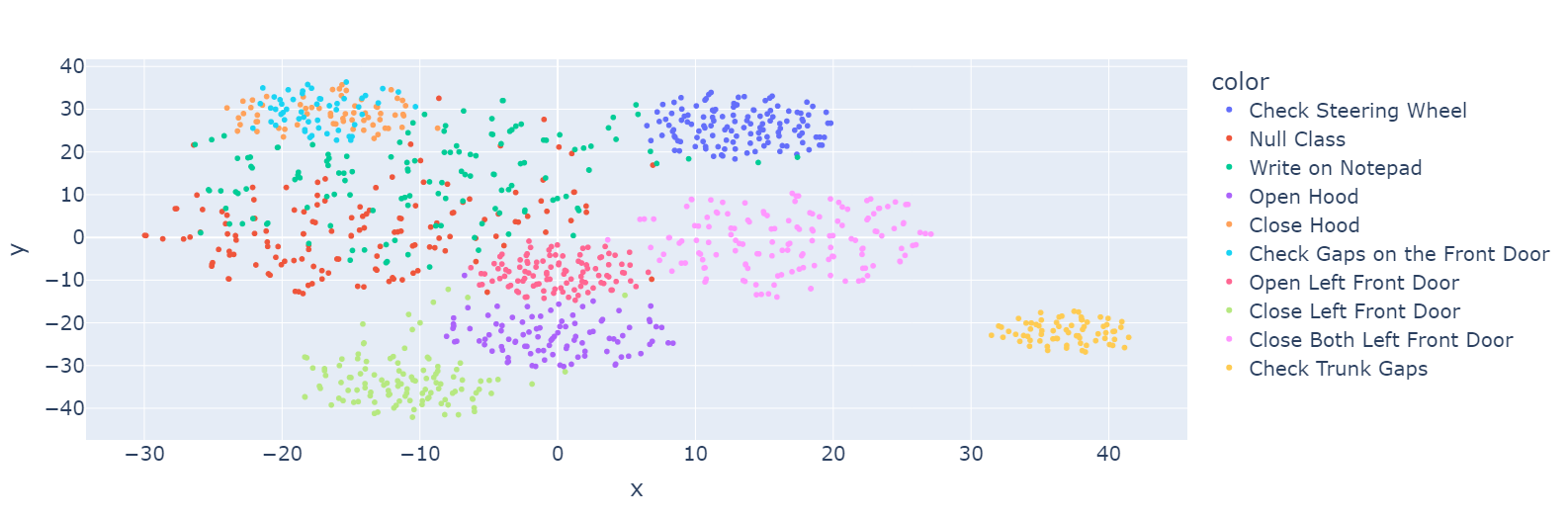}

\subcaption{\label{fig:label} 2D visualization of the proposed MICS-EFS method}
\end{figure}

\end{minipage}
\clearpage

\begin{IEEEbiography}[{\includegraphics[width=1in,height=1.25in,clip,keepaspectratio]{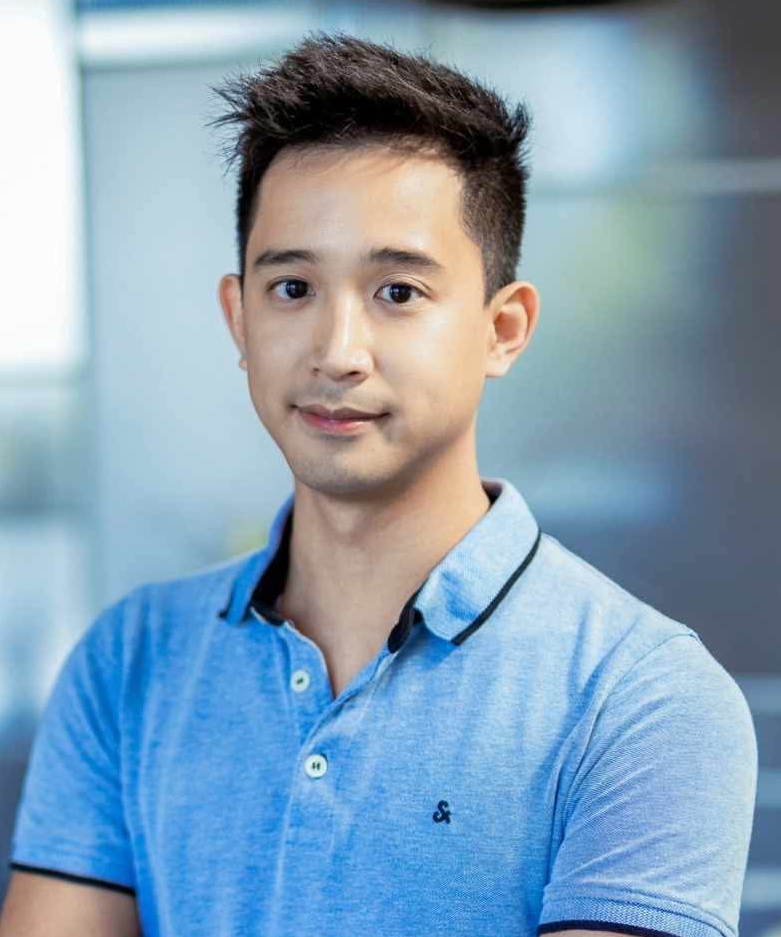}}]{A. T. Hoang} is a dedicated data scientist and computer scientist with a strong passion for applying technology in industrial sector. He is currently a Research Associate at the Institute for Computer Science and Control (SZTAKI) and a member of the Intelligent Process Research Group within the Engineering and Intelligence Management Laboratory. As a Ph.D. student at Eötvös Loránd University (ELTE), he specializes in deep learning for automatic structure identification and feature selection. He holds a double MSc degree in Data Science from KTH Royal Institute of Technology and Eötvös Loránd University, with a minor in Innovation and Entrepreneurship. With extensive experience in deep learning and computer vision, Anh Tuan has demonstrated success in both academia and industry, contributing significantly to advancements in technology.
\end{IEEEbiography}

\begin{IEEEbiography}[{\includegraphics[width=1in,height=1.25in,clip,keepaspectratio]{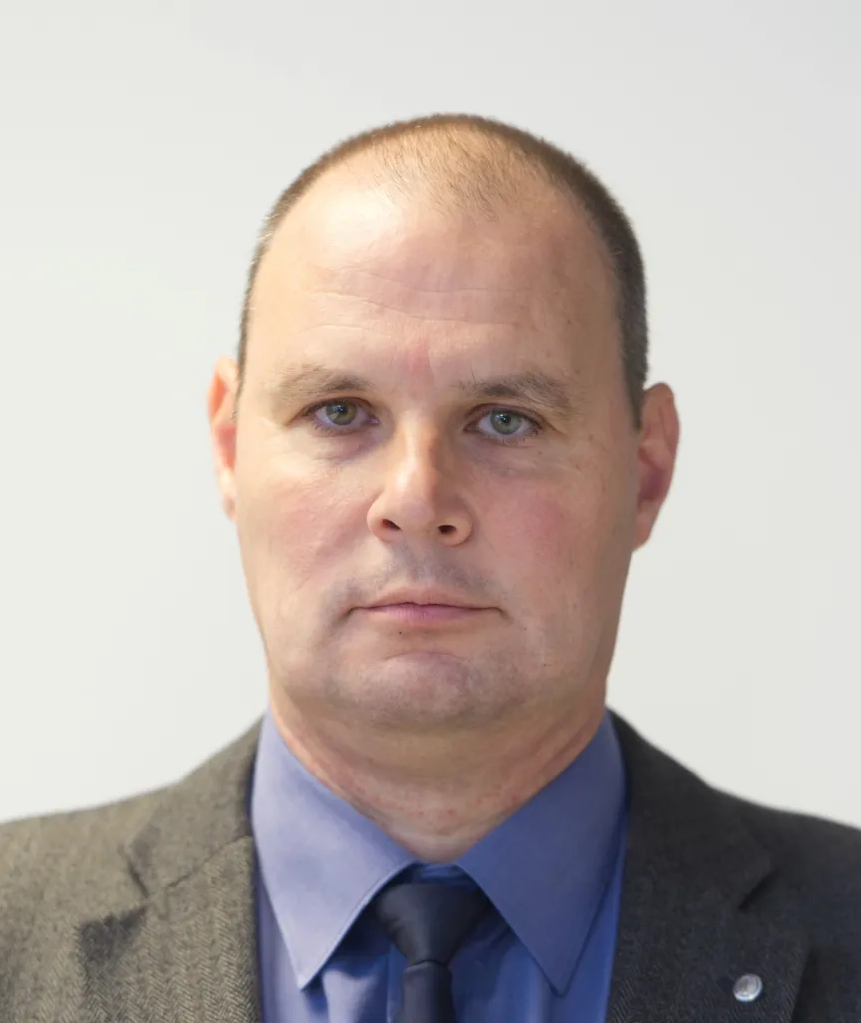}}]{Zs. J. Viharos} MBA, senior research fellow and project manager of the HUN-REN Institute for Computer Science and Control (SZTAKI) and full researcher and lecturer of the John von Neumann University, in Hungary. In the Research Laboratory on Engineering and Management Intelligence at SZTAKI, he leads the Intelligent Processes research group, is supervisor of many PhD programs. He is leading various sizes of industrial, and national or European supported R\&D projects with durations from some months up to many years. His typical roles are project sponsorship, project management and content leadership for industrial projects. He has around 200 scientific publications resulted in around 1000 independent references (2024), is member of the Boards of Reviewers of the scientific journals: Measurement, Measurement: Sensors, Measurement: Food, Applied Intelligence, Reliability Engineering and System Safety and is member, or chair of various scientific conferences. He is Chairperson of the IMEKO TC10 - Measurement for Diagnostics, Optimization and Control and the President of the Hungarian Member Organization of the International Measurement Confederation (IMEKO). He is member of the IEEE (Institute of Electrical and Electronics Engineers), No.: 93787359, and of the International Society of Applied Intelligence, member of the Production Systems section of the Scientific Society for Mechanical Engineering (GTE) in Hungary and member of the Computer and Automation Committee of the public body of the Hungarian Academy of Sciences (MTA), member of the Hungarian Standards Institution (MSZT) and János Bolyai Mathematical Society (in Hungary).
\vspace{6.5cm}
\end{IEEEbiography}



\EOD

\end{document}